\documentclass[10pt]{article} %

\usepackage[preprint]{rlj} %
\usepackage{amssymb}            %
\usepackage{mathtools}          %
\usepackage{mathrsfs}           %
\usepackage{graphicx}           %
\usepackage{subcaption}         %
\usepackage[space]{grffile}     %
\usepackage{url}                %
\usepackage{lipsum}             %
\usepackage{booktabs}           %
\usepackage{wrapfig}            %

\newcommand{\compactlistleftmargin}{1.2em}
\newcommand{\compactlistlabelsep}{0.4em}
\newcommand{\compactlistitemsep}{2pt}
\newcommand{\compactlisttopsep}{2pt}
\newenvironment{compactitemize}{%
  \begin{itemize}[leftmargin=\compactlistleftmargin,labelsep=\compactlistlabelsep,itemsep=\compactlistitemsep,topsep=\compactlisttopsep,parsep=0pt,partopsep=0pt]%
}{\end{itemize}}

\newcommand{\revision}[1]{#1}

\newcounter{question}
\newcommand{\question}[1]{\textbf{Question \refstepcounter{question}\thequestion: #1}}

\newcommand{\figref}[1]{Figure~\ref{#1}}

\newcommand{\blfootnote}[1]{%
  \begingroup
  \renewcommand\thefootnote{}\footnote{#1}%
  \addtocounter{footnote}{-1}%
  \endgroup
}

\title{Task diversity produces systematic transfer\\but inhibits continual reinforcement learning}

\setrunningtitle{Task diversity produces systematic transfer but inhibits continual reinforcement learning}

\author{Purab Seth\textsuperscript{1,2,$*$}, Neil Shah\textsuperscript{1,2,$*$}, Ishaan Sinha\textsuperscript{1,2,$\ddagger$}, \\
Kunal Jha\textsuperscript{3}, Samuel J. Gershman\textsuperscript{1,4}, Max Kleiman-Weiner\textsuperscript{3},\\
Wilka Carvalho\textsuperscript{1,$\dagger$}}

\emails{}

\affiliations{
$^{1}${Kempner Institute for the Study of Natural and Artificial Intelligence, Harvard University}\\
$^{2}${Harvard College}\\
$^{3}${University of Washington}\\
$^{4}${Department of Psychology and Center for Brain Science, Harvard University}
}

\contribution{}{}


\summary{}

\begin{document}

\maketitle  %
\blfootnote{$^{*}$Equal contribution.\quad $^{\ddagger}$Core contributor.\quad $^{\dagger}$Corresponding author: \texttt{wilka@humanagencylab.io}.}

\begin{abstract}
\revision{
Continual reinforcement learning (RL) aims to produce agents that never stop adapting to new tasks.
A key question is how this interacts with the diversity of tasks an agent experiences.
Prior work has shown that training on many diverse tasks leads to agents with strong zero-shot and in-context adaptation.
However, this work evaluated agents \textit{after} they'd stopped learning, i.e. with frozen weights.
How task diversity affects an agent's ability to \textit{continue} learning over a sequence of distribution shifts remains unclear. %
}%
We introduce Banyan, a GPU-accelerated continual RL domain where one can parametrically control three independent axes that define a task: the map layouts an agent must navigate, the objects it must interact with, and the hierarchical structures of sub-goal dependencies.
We find that increasing diversity along each axis induces \textit{systematic transfer}---that is, agents begin training on a new task distribution near the performance attained on the previous one, even when the shift changes the structure of the optimal policy.
While increasing diversity improves systematic transfer, we find that \textit{too much} diversity inhibits a learner's ability to continue adapting to new task distributions.
As diversity increases, learners plateau in the success rate they achieve on new tasks, yet continue improving on old tasks---even without further exposure to them. %
\revision{
We find this phenomenon manifests across continual learning algorithms, memory architectures, architecture sizes, and in Kinetix---a physics-based control domain.
We release Banyan as a domain for running controlled experiments that study continual RL in the many-tasks regime}. Code is available at \url{https://github.com/nhshah15/banyan}.
\end{abstract}

\section{Introduction}

\revision{What do we need to develop reinforcement learning (RL) agents that can continuously generalize to new tasks? For much of the history of RL, researchers have focused on developing novel agent components for promoting generalization\citep{mazzaglia2024genrl, oh2017zeroshot, klissarov2025maestromotif, successor-transfer, carvalho2023composing, carvalho2023combining}. Recent work, however, has shown that task diversity itself---paired with basic but powerful learning algorithms and architectures---can promote remarkable generalization. Agents trained on many diverse tasks achieve zero-shot sim-to-real transfer \citep{domain-randomization}, generalize zero-shot to held-out games built from the same task primitives \citep{xland, xland-minigrid}, and generalize to human partners without any human training data \citep{cec}. These works, however, have primarily studied generalization while an agent's weights are held fixed.}

\revision{One prominent idea is that task diversity can induce a generalizable learning strategy. This is perhaps best displayed by work showing that training on diverse tasks gives rise to in-context learning, a form of meta-learning in which an agent uses the examples it sees in context to generalize to new tasks within that distribution \citep{ada, kirsch2022general}. However, this work, too, has focused primarily on evaluating agents with fixed weights. What happens when the weights are allowed to change and the task distribution shifts many times?}

\revision{In this work, we study how task diversity affects continual RL, where an agent must continuously generalize to new tasks \citep{crl-def}. Prior work has commonly studied adaptation to a new task after an agent has converged on the current one \citep{kirkpatrick2017overcoming, rusu2016progressive, ahn2025reset}. However, the world does not wait for an agent to converge. On some tasks an agent will have converged, on others not.} The question is therefore not only whether a final policy generalizes, but whether the learning process itself remains adaptive under task distribution shifts.

\revision{Existing environments have not allowed researchers to run controlled experiments on how task diversity affects generalization as an agent continually transfers to new task distributions. Continual RL benchmarks study sequential learning but typically contain at most hundreds of tasks \citep{continual-world, coom}. GPU-accelerated environments with procedurally generated tasks contain millions or billions of tasks \citep{xland-minigrid, craftax}, but are not designed to isolate how the amount and type of diversity affect continual learning across controlled distribution shifts.}

\revision{To enable researchers to run controlled experiments studying how task diversity affects continual RL}, we introduce Banyan, a GPU-accelerated domain for continual RL over compositional task spaces. A task is specified by a task tree and a layout (see \figref{fig:env}). Task trees specify which objects must be combined or transformed, and in what order, to produce a goal object. Layouts govern how the agent navigates the map. This factorization yields three axes of controllable diversity: layouts, task-tree topologies, and task-tree instances. Banyan supports procedural generation along each axis independently, yielding billions of possible compositional tasks while preserving control over the source and amount of diversity.

Using Banyan, we study whether task diversity produces learning that transfers across distribution shifts, and whether this transfer compounds into continual learning. The answer is mixed.
Our contributions are as follows:
\begin{compactitemize}
    \item \textbf{Banyan, a controlled many-task domain for continual RL.}
    We introduce a GPU-accelerated procedurally generated environment in which layouts, task-tree instances, and task-tree topologies can be varied independently. This makes it possible to (1) isolate how the amount and type of diversity affect learning under distribution shift, and (2) study continual RL over billions of tasks.

     \item \textbf{Experimental evidence that task diversity improves forward and backward transfer across task distributions.}
    When agents train on one task distribution and then continue on a different distribution, increasing diversity along each axis reduces the performance drop at the shift boundary.

    \item \textbf{Experimental evidence that diversity does not by itself enable longer-run continual learning, but can inhibit it.}
    When agents train on a small amount of diversity, we see that this improves learning on a sequence of new task distributions; however, large amounts of diversity hinder learning on new distributions. \revision{We find that this phenomenon occurs in other domains, across different architectures, in large architectures, and with different continual learning algorithms.}
\end{compactitemize}

\section{Related work}

A key question in reinforcement learning is how experience on one set of tasks affects performance and learning on another. \textbf{Forward transfer} measures whether prior experience improves initial performance or learning speed on new tasks; \textbf{backward transfer} measures how later learning changes performance on previously encountered tasks, with negative backward transfer corresponding to \textbf{forgetting}.
Much of the transfer literature builds mechanisms that make experience reusable: successor features factor value functions into reusable dynamics-dependent features and task-specific reward weights \citep{successor-transfer}, hierarchical methods compose lower-level skills into higher-level behavior \citep{temporal-abstraction, maxq-rl, option-critic}, and multitask and meta-learning methods learn policies, representations, or initializations that can be adapted to new tasks \citep{distral, maml}. These approaches ask how agents should be structured so that experience transfers. We ask the complementary question: to what extent can transfer be induced by the structure of the task distribution itself?

\textbf{Task diversity as a source of transfer.}
A growing body of work shows that diversity in the training distribution can itself drive generalization. Domain randomization varies visual or physical properties to enable sim-to-real transfer \citep{domain-randomization, dynamics-randomization, rubiks-cube}; procedural generation constructs distinct training and test distributions over levels or tasks \citep{procgen-generalization, procgen}; and large-scale environments such as XLand and AdA show that broad task distributions can support zero-shot generalization and in-context adaptation \citep{xland, ada}. Recent work further shows that task diversity can improve generalization under fixed data budgets \citep{impala-atari} and support coordination with novel partners \citep{cec}. Together, these results establish task diversity as a powerful source of post-training generalization and adaptation. They do not, however, show whether diversity makes the learning process itself more transferable as agents continue updating their weights under distribution shift.

\textbf{Systematic transfer.}
 In compositional domains, the strongest form of transfer is not merely robustness to new instances, but systematic reuse of familiar components in new combinations. This capability is often studied as \textbf{systematic generalization} in cognitive science and machine learning \citep{fodor-connectionism, lake-generalization, bahdanau-systematic}. In RL, related questions arise in modular policy learning, compositional task specifications, and compositional control benchmarks \citep{andreas-modular, nangue-boolean, mendez-lifelong-learning, mendez-2022-lifelong-learning}. None of this work studied whether the data distribution itself could give rise to systematic generalization.
Most similar to our work are \citet{hill2019environmental} and \citet{lake2023human}, who both study how an agent's data distribution can give rise to systematic generalization. However, neither studied this within a continual learning setting, and \citet{lake2023human} focused on natural language processing tasks, not RL tasks.

\textbf{Many-task environments for continual RL.}
Studying how task diversity affects continual learning requires both scale and control. Continual RL benchmarks make sequential learning explicit and measure transfer and forgetting across changing tasks \citep{continual-world, coom, meal}, but they typically contain at most hundreds of tasks. GPU-accelerated procedurally generated environments make many-task training practical \citep{xland-minigrid, craftax}, but they are not designed and have not been used to independently vary the amount and type of diversity across controlled distribution shifts. Banyan fills this gap by providing a GPU-accelerated continual RL domain in which layouts, task-tree instances, and task-tree topologies can be varied independently.

\section{\revision{Banyan: a domain for controlled continual RL experiments}}
\label{sec:domain}

\begin{figure}[h]
    \centering
    \includegraphics[width=0.95\linewidth]{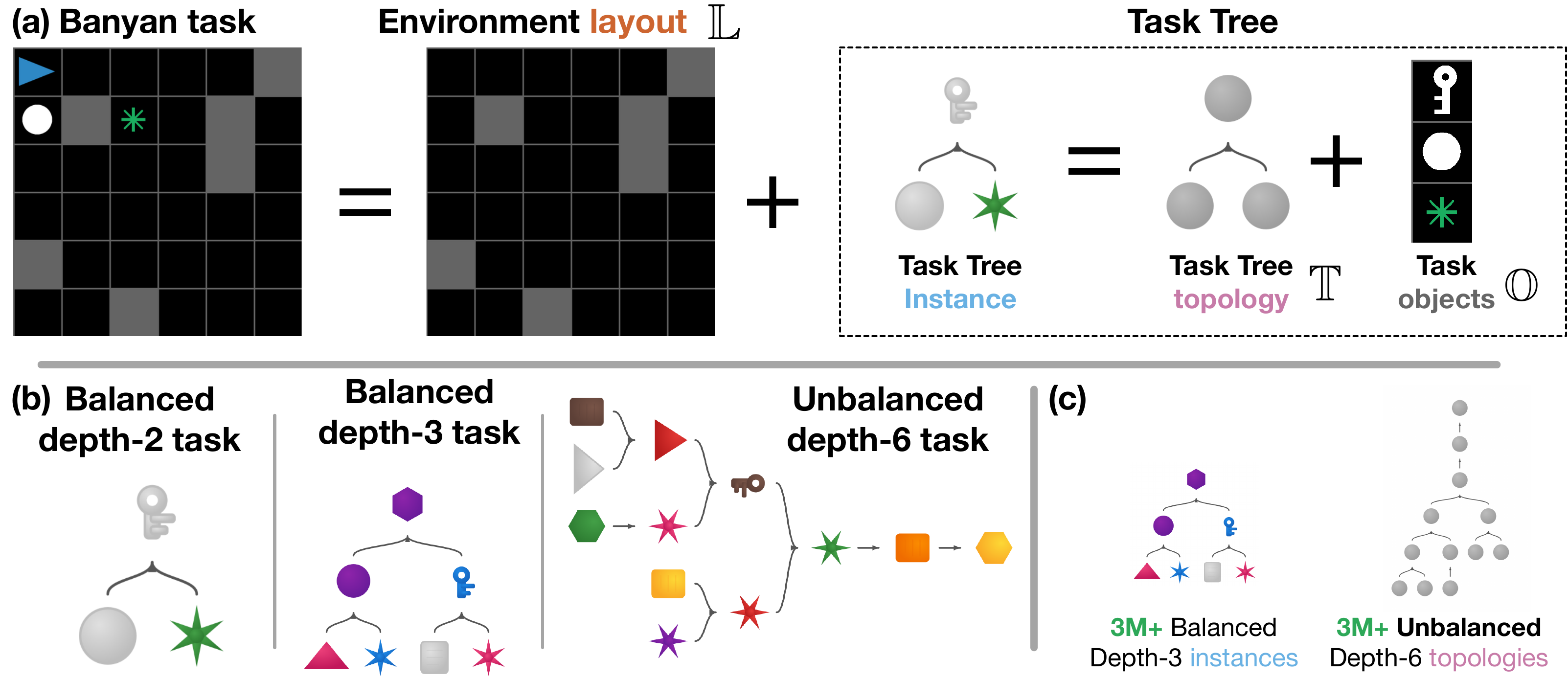}
     \caption{Overview of Banyan. (a) A task decomposes into an environment layout and a task tree; each task tree further decomposes into a topology and an instance over task objects. (b) Tree depth and topology control the horizon and dependency structure of a task. (c) Procedural generation yields millions of task instances and topologies, enabling controlled many-task continual RL.}
    \label{fig:env}
\end{figure}

\newcommand{\task}{\tau}
\newcommand{\layout}{\mathbb{L}}
\newcommand{\objects}{\mathbb{O}}
\newcommand{\topology}{\mathbb{T}}
Banyan is a MiniGrid-style \citep{minigrid} domain for studying continual RL over large compositional task spaces. A Banyan task $\tau = \langle \layout, \topology, \objects \rangle$ has three independently varied parts (Figure~\ref{fig:env}): a layout $\layout$, a task-tree topology $\topology$, and an object assignment $\objects$. The \textbf{layout} fixes the wall configuration and the initial placement of objects, and so determines the navigation problem the agent must solve. The remaining two pieces specify a \textbf{task tree}: a compositional recipe for the goal object. Its leaves are the objects placed in the layout; each internal node is an object produced by transforming or combining its children; the root is the goal. The depth of the tree is the longest leaf-to-root path, so deeper tasks require longer sequences of object interactions. Within the task tree, the \textbf{topology} $\topology$ is the abstract shape --- depth, branching, and which nodes are transformations versus combinations --- and the \textbf{object assignment} $\objects$ pins concrete identities to the nodes of that shape. Changing $\topology$ alters which sub-goals must be composed and in what order, and so changes the optimal subgoal policy. Changing $\objects$ preserves the solution structure but changes which objects ground it.

\textbf{Interacting with Banyan.}
When an episode begins, a layout and task tree are sampled. First, the layout walls are placed; afterwards, the leaf nodes associated with the task tree are placed.
\textbf{Observations}. The agent observes the entire grid and a description of the goal object (the root node of the task tree). It does not observe the task tree structure and must learn which sequences of object interactions complete a task.
\textbf{Actions}. The agent can move around the map using the ``up'', ``down'', ``left'', ``right'', and ``stay'' actions. The agent can use the ``pickup''  and ``drop'' actions to move items around the map and place particular items adjacent to one another to combine them and create a new item. The agent can also use the ``toggle'' action while standing on a tile with an item to perform a single-item transformation that creates a new item when needed.
\textbf{Dynamics}. How two objects combine or the effect of ``transform'' on a single object stays fixed across tasks. Tasks therefore differ in where the agent must move, which objects are needed, and how those objects compose into the goal.
\textbf{Rewards}. The agent receives a terminal success reward of $+1$ when the root object is produced and a penalty of $-1$ if a dead-end condition is triggered. Dead-ends are triggered when the agent places two items adjacent to one another that combine under the world's merge rules but whose combination is not part of the current task tree, consuming an object the task requires. 
\textbf{Episode termination}. An episode terminates if (i) the root goal is achieved, (ii) the episode hits the max-step horizon (timeout), or (iii) a dead-end condition is triggered.
We describe the domain in more detail in \S\ref{subsec:domains-banyan}.

\textbf{How task trees relate to their optimal solutions.}
Holding the topology fixed while sampling new instances creates many tasks with the same solution structure but different object identities. Holding the object rules fixed while sampling new topologies creates new subgoal dependencies from familiar components. Increasing depth changes the horizon of the problem, creating long dependency chains that are harder to solve than shallow trees.

\textbf{How changing a layout, instance, or topology affects transfer.}
Layout shifts vary navigation while preserving the object-composition problem. Instance shifts vary the objects grounding a task while preserving the abstract task structure. Topology shifts vary the task structure itself, making them the strongest test of whether learning transfers compositionally.

\section{Experiments}
\label{sec:experiments}

\revision{Does task diversity induce a generalizable learning strategy that keeps promoting transfer across multiple distribution shifts?} We address this with two sets of experiments. \S\ref{sec:two-dist} studies a single shift: how does diversity on the first distribution shape forward transfer to the second, and backward transfer to the first once training completes? \S\ref{sec:ten-dist} extends the protocol to a sequence of nine shifts, and asks whether the single-shift effect of diversity persists as shifts compound.

A \textbf{task distribution} $d$ is a distribution over Banyan tasks $\tau = (\layout, \topology, \objects)$. Every distribution we use spans depths $1$ through $6$, and we require the depths to nest: every task at depth $l$ also appears as a sub-task inside some task at every deeper level. (As a reminder, a tree's depth is the length of the longest leaf-to-root path; see \figref{fig:env} for examples). The constraint serves two purposes. Deeper tasks are hard to learn from scratch, so the shallower depths act as a built-in curriculum; and the nesting makes diversity easy to count: the depth-$6$ tasks determine everything shallower, so we count diversity by the depth-$6$ tasks alone. A distribution with $n$ topologies has $6n$ tasks in total but effective diversity $n$.

\textbf{Metrics}. To study how task diversity affects transfer, we use metrics for forward and backward transfer between task distributions. For \textbf{forward transfer}, we define the \textbf{transfer gap} between adjacent distributions: the difference between the agent's terminal success rate on one distribution, $S_{\text{end}}(d_i)$, and its initial success rate on the next, $S_{\text{start}}(d_{i+1})$:
\begin{equation}
    \Delta_i = S_{\text{start}}(d_{i}) - S_{\text{end}}(d_{i-1}) .
\end{equation}
Negative values mean success drops at the shift; a value near zero means the agent begins $d_i$ at the level it reached on $d_{i-1}$. For \textbf{backward transfer}, we ask how later training changes performance on an earlier distribution. Let $S_{\text{end}}(d_i; k)$ denote success on $d_i$ evaluated after training through phase $k$, with $S_{\text{end}}(d_i; i) = S_{\text{end}}(d_i)$. Backward transfer is the change in success on $d_i$ between the end of its own phase and the end of phase $k$:
\begin{equation}
    B(k, i) = S_{\text{end}}(d_i; k) - S_{\text{end}}(d_i; i).
\end{equation}
Positive values mean later training improved performance on $d_i$; negative values mean forgetting.
All plots show mean and standard error across seeds.

\textbf{Baselines}. This work seeks to understand how data diversity can enable transfer and continual learning in RL algorithms. We thus focus on algorithms that minimally use auxiliary tasks and have been shown to benefit from data diversity. For policy-based methods, we focus on Proximal Policy Optimization~\citep[PPO;][]{schulman2017proximal}. For value-based methods, we focus on Parallelised Q-Network~\citep[PQN;][]{Gallici25simplifying}. See \S\ref{sec:appendix-algorithms} for a summary of the methods we study.

\subsection{Diversity produces systematic transfer}
\label{sec:two-dist}
\begin{figure}[h]
    \centering
    \includegraphics[width=0.95\linewidth]{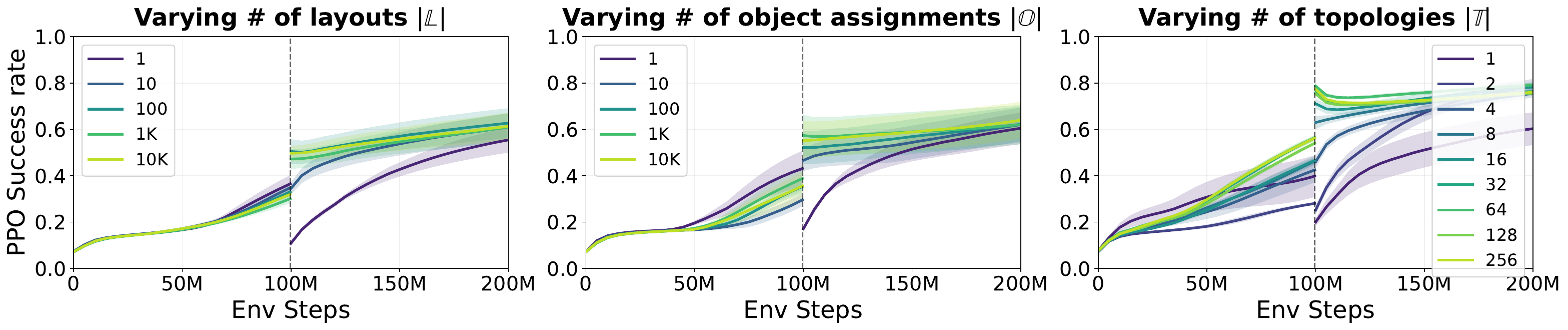}\\[2pt]
    \includegraphics[width=0.95\linewidth]{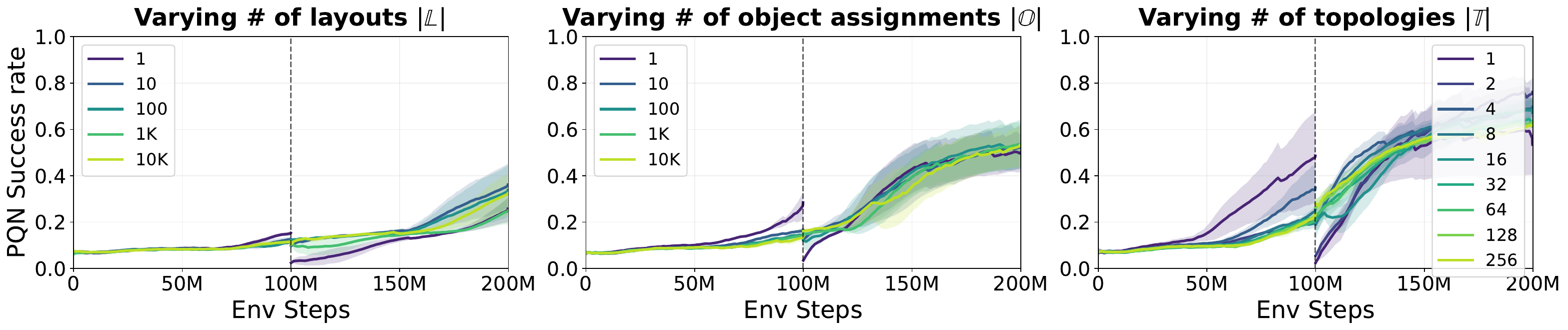}
    \caption{
\textbf{Increasing task diversity in $d_1$ leads to the transfer gap $\Delta_2$ to approach 0}. Increasing the number of layouts $\layout$, object assignments $\objects$, or topologies $\topology$, leads agents to begin $d_2$ close to their performance on $d_1$. This is true for both PPO (top) and PQN (bottom). Dashed lines mark the $d_1 \to d_2$ boundary.
}
\label{fig:r1}
\end{figure}

\textbf{Experimental setup.} We test how holding the world fixed and varying diversity along a single axis affects transfer. In $d_1$ we vary the number of layouts $\layout$, task-tree object assignments $\objects$, or task-tree topologies $\topology$; $d_2$ contains 10K new held-out items along the same axis for layouts and object assignments, or $256$ for topologies. \revision{Each PPO run has 10 seeds and each PQN run has 3 seeds.}
\question{How does increasing task diversity in $d_1$ affect forward transfer to $d_2$?}
\revision{Across all dimensions we vary, increasing the number of examples in $d_1$ causes $\Delta_2$ to approach zero (Figure~\ref{fig:r1}; Figure~\ref{fig:appendix-transfer-gap}).}
Layout count $|\layout|$ and object assignment count $|\objects|$ qualitatively have the same effect on reducing the transfer gap, whereas topology count $|\topology|$ seems to have a different effect: learning seems to benefit strongly from increasing topology count.
We attribute this asymmetry to what each axis demands of the policy.
Ignoring efficiency, a policy that is optimal for one layout $\layout$ or task-tree object assignment $\objects$ of a given topology $\topology$ can reapply the subtask progression order it has learned to other layouts or task-tree object assignments of that topology. In contrast, every topology requires its own progression order, so spanning the family requires learning a richer set of knowledge.
The results hold for both PPO and PQN, which suggests the effect is not an artifact of how policy and value functions absorb structure.

\begin{figure}
    \centering
    \includegraphics[width=0.95\linewidth]{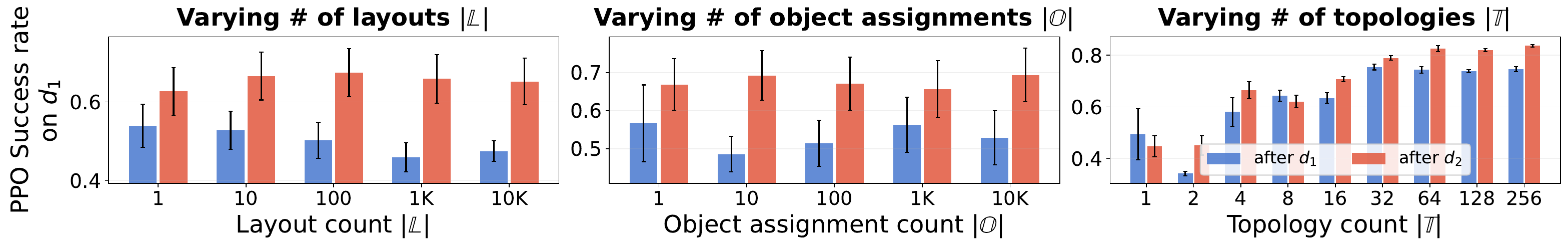}\\[2pt]
    \includegraphics[width=0.95\linewidth]{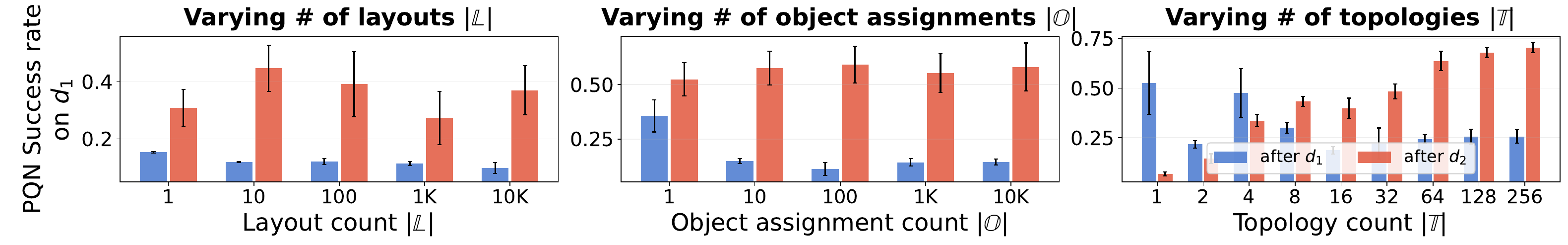}
    \caption{
\textbf{Increasing task diversity in $d_1$ improves backward transfer to $d_1$ after $d_2$, but only when varying topology count $|\topology|$}.
    }
    \label{fig:r2}
\end{figure}

\question{How does increasing task diversity affect backward transfer to $d_1$ after the agent has been exposed to $d_2$?}
Neither layout count $|\layout|$ nor object assignment count $|\objects|$ visibly affects backward transfer across PPO and PQN. However, varying topology count $|\topology|$ leads to markedly better backward transfer across both PPO and PQN (\figref{fig:r2}; \figref{fig:appendix-b-independent}). As with the forward-transfer result, we find that varying layout count and varying object assignment count have a similar effect that is qualitatively different from varying topology count.

\begin{figure}
    \centering
    \includegraphics[width=0.95\linewidth]{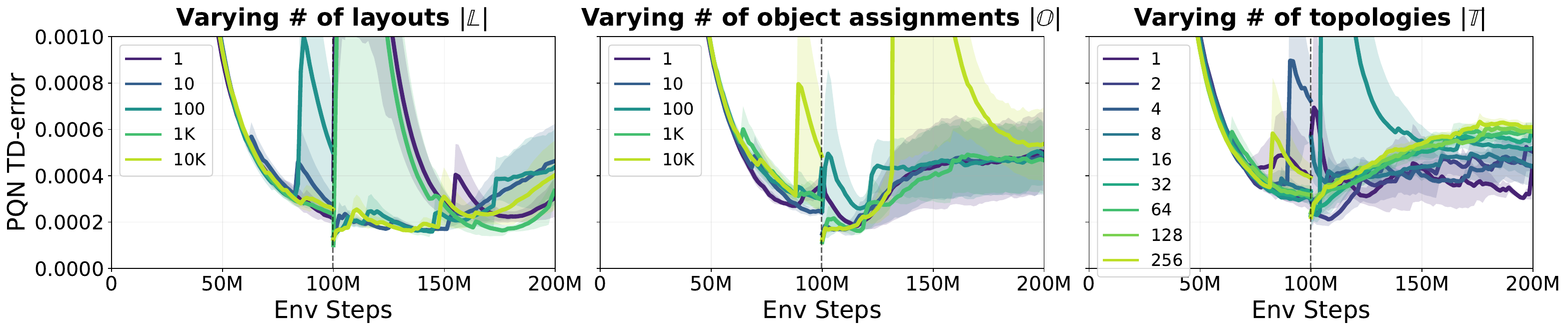}
    \caption{
        \textbf{Increasing task diversity in $d_1$ leads to value-predictions in $d_2$ with a lower initial TD-error}. Dashed lines mark the $d_1 \to d_2$ boundary.
        }
        \label{fig:pqn-td}
    \end{figure}
\question{Why does increasing task diversity induce systematic transfer?}
One simple hypothesis is that diversity produces better representations, and better representations produce better predictions. PQN lets us check this directly: its policy is derived from its value prediction, so prediction quality is observable as TD-error. We examine the TD-error at the $d_1 \to d_2$ boundary --- the first $d_2$ data the agent's $d_1$-trained predictions are evaluated on. \revision{Higher diversity along every axis yields both a more performant initial policy (Figure~\ref{fig:r1}) \textit{and} a smaller initial TD-error at the onset of $d_2$ (Figure~\ref{fig:pqn-td}).} As is a common empirical phenomenon with value-based methods, as PQN improves, the magnitudes of both predicted return and TD-error grow together.

\begin{wrapfigure}[11]{r}{0.65\linewidth}
    \centering
    \vspace{-15pt}
    \includegraphics[width=0.49\linewidth]{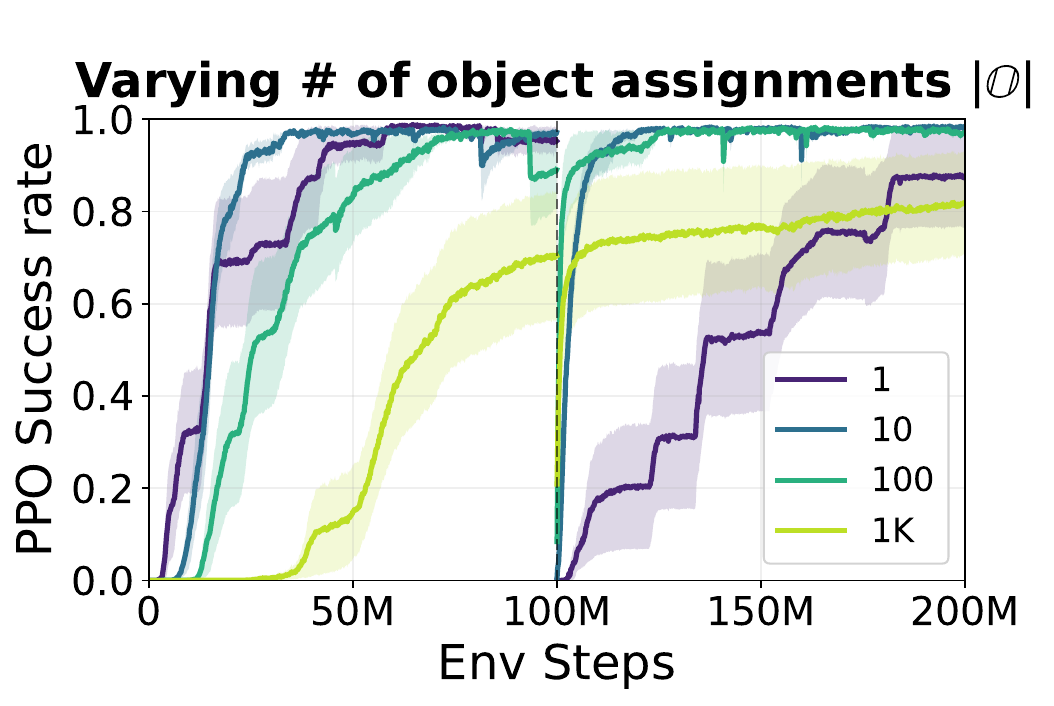}\hfill
    \includegraphics[width=0.49\linewidth]{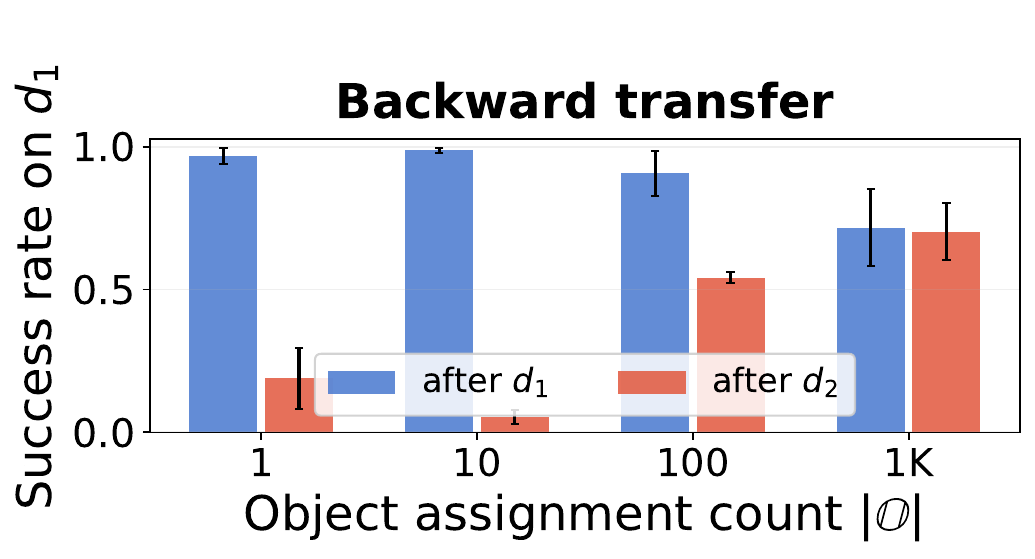}
    \caption{
        \textbf{Task diversity also improves forward and backward transfer in a continuous control domain.}
        }
    \label{fig:continuous-banyan}
\end{wrapfigure}
\question{Do the effects of task diversity on transfer hold in a different control task?}
We adapt PointMass~\citep{connell1992sss,sutton2018}, a domain in which an agent controls a 2D point mass to move objects around by bumping into them, to support tree-structured tasks.
\textbf{Actions}. The agent selects a continuous $(x,y)$ direction to move one unit per step and moves objects by hitting them. Two objects can be combined by placing both inside a special ``combine'' object, which then merges them according to the task tree (See \S\ref{subsec:domains-pointmass} for more details). We run PPO only, as value-based methods require additional care to extend to continuous action spaces, which we leave for future work. We sweep the object assignment axis $\objects$ from $1$ to $1{,}000$ balanced trees at depths $1$ and $2$. Like before, we find that diversity improves systematic transfer---$\Delta_2$ rises from $\approx -0.95$ at $n=1$ to $\approx -0.5$ at $n=1000$---and improves backward transfer---$B(2,1)$ rises from $\approx -0.8$ to $\approx 0$ across the same range (\figref{fig:continuous-banyan}; \figref{fig:appendix-b-pointmass}).

\subsection{Systematic transfer does not enable continual reinforcement learning}
\label{sec:ten-dist}

\revision{We just showed that increasing task diversity can improve both forward and backward transfer within a single distribution shift. Here we ask whether this effect compounds across longer sequences by studying transfer across ten task distributions. We focus on PPO, which we found more performant than PQN in the continual setting. All experiments use $3$ seeds unless otherwise stated.}

\begin{figure}[h]
    \centering
    \includegraphics[width=.9\linewidth]{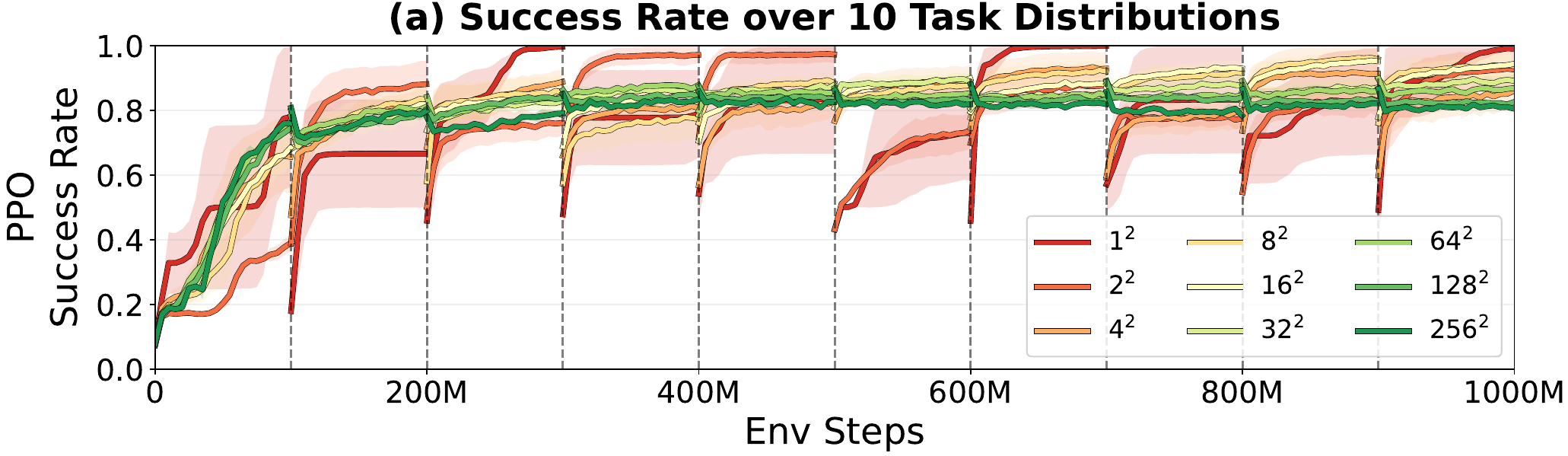}\\[2pt]
    \includegraphics[width=.9\linewidth]{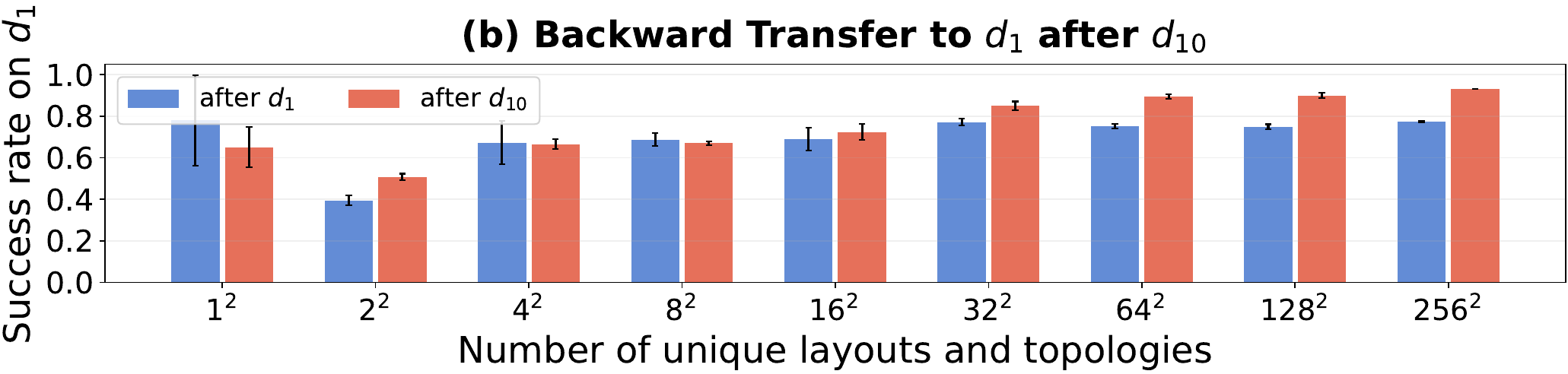}
    \caption{\revision{
        \textbf{Task diversity inhibits forward transfer in continual RL but improves backward transfer}.
        (a) Task diversity leads to continual improvement at intermediate diversity ($n^2 = 16^2$), but inhibits continual learning at high diversity ($n^2 = 256^2$).
        Success rates across 9 distribution shifts, averaged over depths 1--6.
        (b) Increasing task diversity improves backward transfer to the first task distribution after 9 distribution shifts.
        Dashed lines mark the $d_i \to d_{i+1}$ boundary.}}
    \label{fig:ten-dist-top}
\end{figure}
\question{How does task diversity affect transfer across a sequence of 10 task distributions?}
We co-vary layout and topology diversity from $n=1$ to $n=256$, yielding up to $n^2=65{,}536$ depth-6 layout--topology combinations per distribution. As in the single-shift experiment, increasing diversity drives $\Delta_i \to 0$ at every transition: at high $n$, each new distribution begins at the success rate the agent reached on the previous one (Figure~\ref{fig:ten-dist-top}).
\revision{At intermediate diversity ($n=16$) the agent continues to learn after each transfer: $S_{\text{end}}(d_i)$ climbs from $\approx 0.69$ at $d_1$ to $\approx 0.94$ at $d_{10}$.}
\revision{At the highest diversity, this intra-distribution improvement shrinks: at $n=256$, $S_{\text{end}}(d_i)$ moves only from $\approx 0.79$ at $d_3$ to $\approx 0.81$ at $d_{10}$.}
 The endpoint $S_{\text{end}}(d_{10})$ in fact declines monotonically as diversity grows past $n=16$, \revision{falling from $0.94$ at $n=16$ through $0.89, 0.87, 0.83$ to $0.81$ at $n=256$.}
Backward transfer moves in the opposite direction: as diversity grows, the agent's performance on $d_1$ after the full sequence exceeds its level at the end of phase 1; \revision{at $n=256$, $S_{\text{end}}(d_1; 1) \approx 0.77$ rises to $S_{\text{end}}(d_1; 10) \approx 0.93$} (Figure~\ref{fig:ten-dist-top}; \figref{fig:appendix-b-ppo}).
This suggests that when an agent is forced to shift task distributions, the network keeps learning the general task structure but is inhibited in its ability to learn the structure specific to each new task distribution.
\revision{These results hold as we increase the capacity of the network (\figref{fig:netsize}).}

\begin{figure}[t]
    \centering
    \includegraphics[width=.9\linewidth]{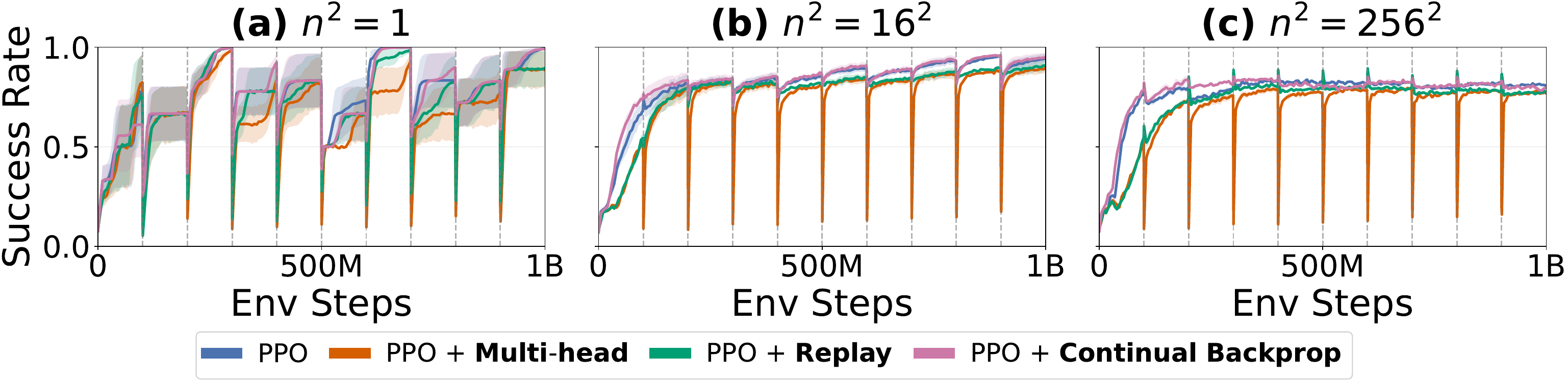}\\[2pt]
    \includegraphics[width=.9\linewidth]{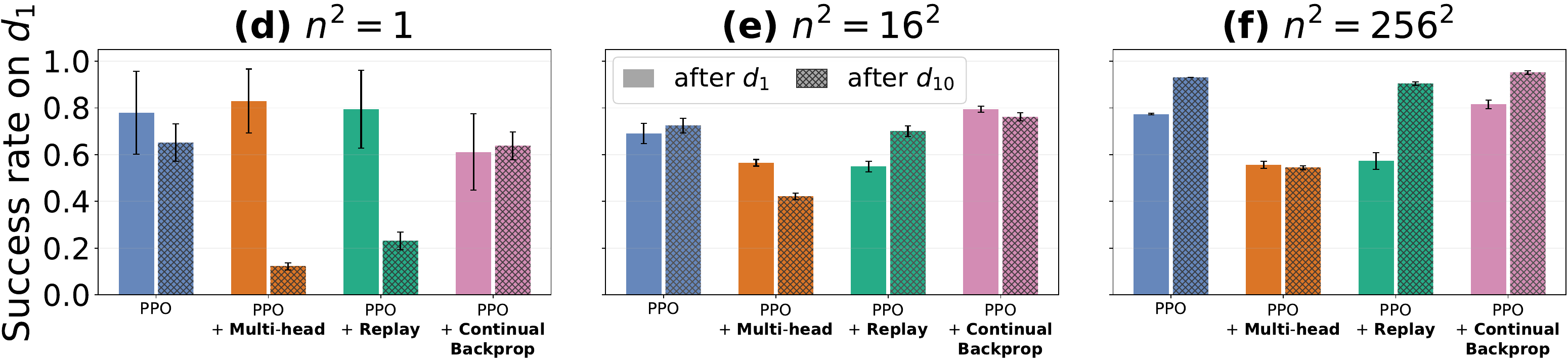}
    \caption{\revision{\textbf{No continual-learning approach mitigates the negative effect of task diversity on continual RL, but methods with a shared policy head retain the backward-transfer benefit.}}}
    \label{fig:methods}
    \vspace{-15pt}
\end{figure}
\revision{\question{How do different continual learning methods better accommodate task diversity during continual RL?}}
\revision{We study three continual learning methods:
(1) \textbf{Continual backprop}~\citep{dohare2024loss}, which targets the loss of plasticity that arises under repeated distribution shifts;
(2) \textbf{Replay}~\citep{shin2017generative}, which continues training on earlier task distributions so the agent never trains on the current distribution alone;
(3) \textbf{Multi-head} policies~\citep{kessler2022same, caccia2023taskagnostic}, which give each distribution its own output head while a shared body absorbs the structure the distributions have in common. No method restores continued improvement at high diversity (\figref{fig:methods}, (a--c)). Unsurprisingly, since the multi-head policy must learn a new output head for each distribution, its success drops to about $0.1$ at each shift before recovering to the level of the other methods. For backward transfer, every method with a shared policy head retains the benefit of diversity. Multi-head policies, however, lose it entirely (\figref{fig:methods}, (d--f); \figref{fig:appendix-b-methods}). This suggests that sharing a policy head is key to learning shared structure across task distributions.}

\begin{figure}[h]
    \centering
    \includegraphics[width=.9\linewidth]{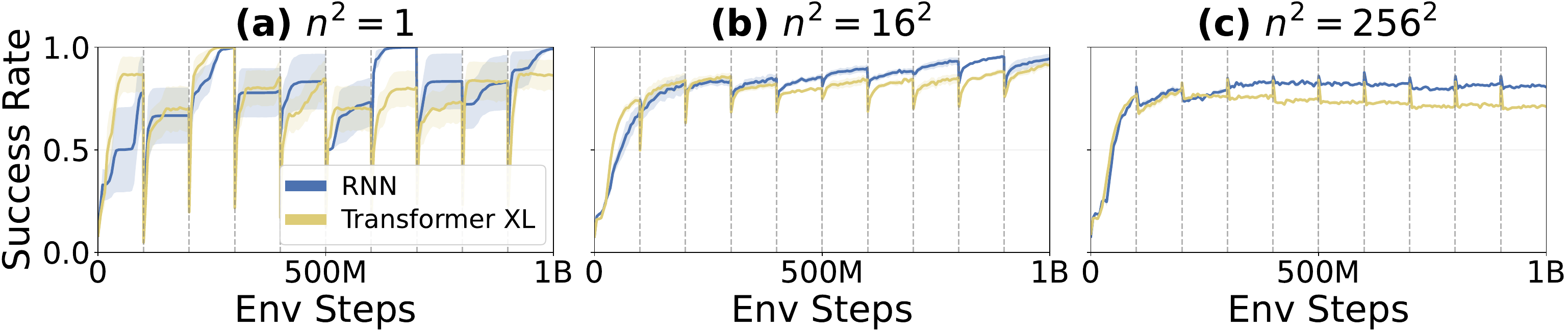}\\[2pt]
    \includegraphics[width=.9\linewidth]{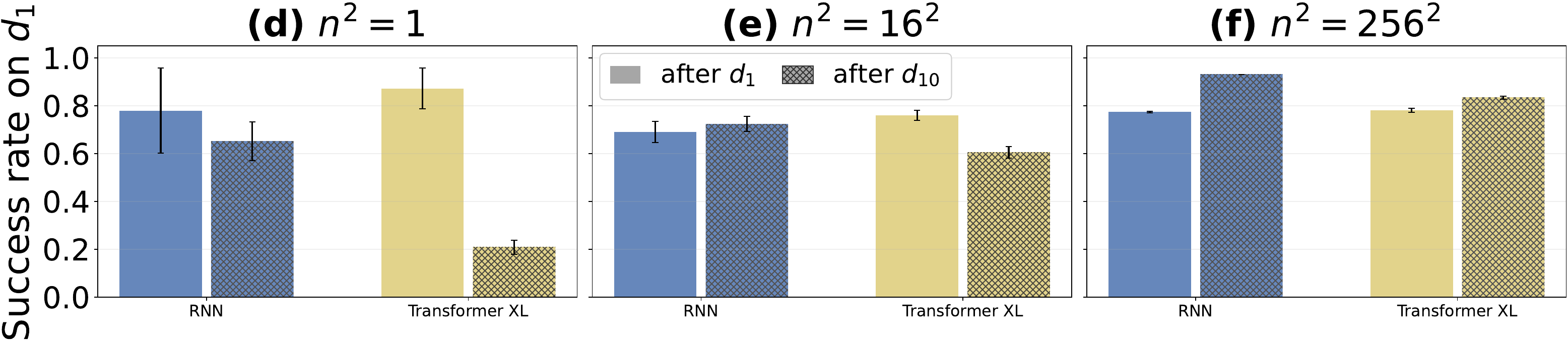}
    \caption{\revision{\textbf{Using a Transformer memory architecture does not mitigate the negative effect of task diversity on continual RL.} \revision{$5$ seeds for the Transformer-XL.}
    }}
    \label{fig:transformer}
\end{figure}

\revision{\question{Do transformers better accommodate task diversity during continual RL?}}
\revision{The plateau at high diversity could reflect the recurrent network rather than the learning problem, so we ask whether a transformer absorbs high-diversity task structure that an RNN cannot. We adapt an open-source implementation of PPO + Transformer-XL~\citep{parisotto2020stabilizing} and hold everything else fixed. Transformers reproduce the same qualitative forward and backward transfer (\figref{fig:transformer}; \figref{fig:appendix-b-transformer}): at high diversity, continued improvement stalls and backward transfer improves with diversity, just as for the RNN.}

\begin{figure}[h]
    \centering
    \includegraphics[width=.7\linewidth]{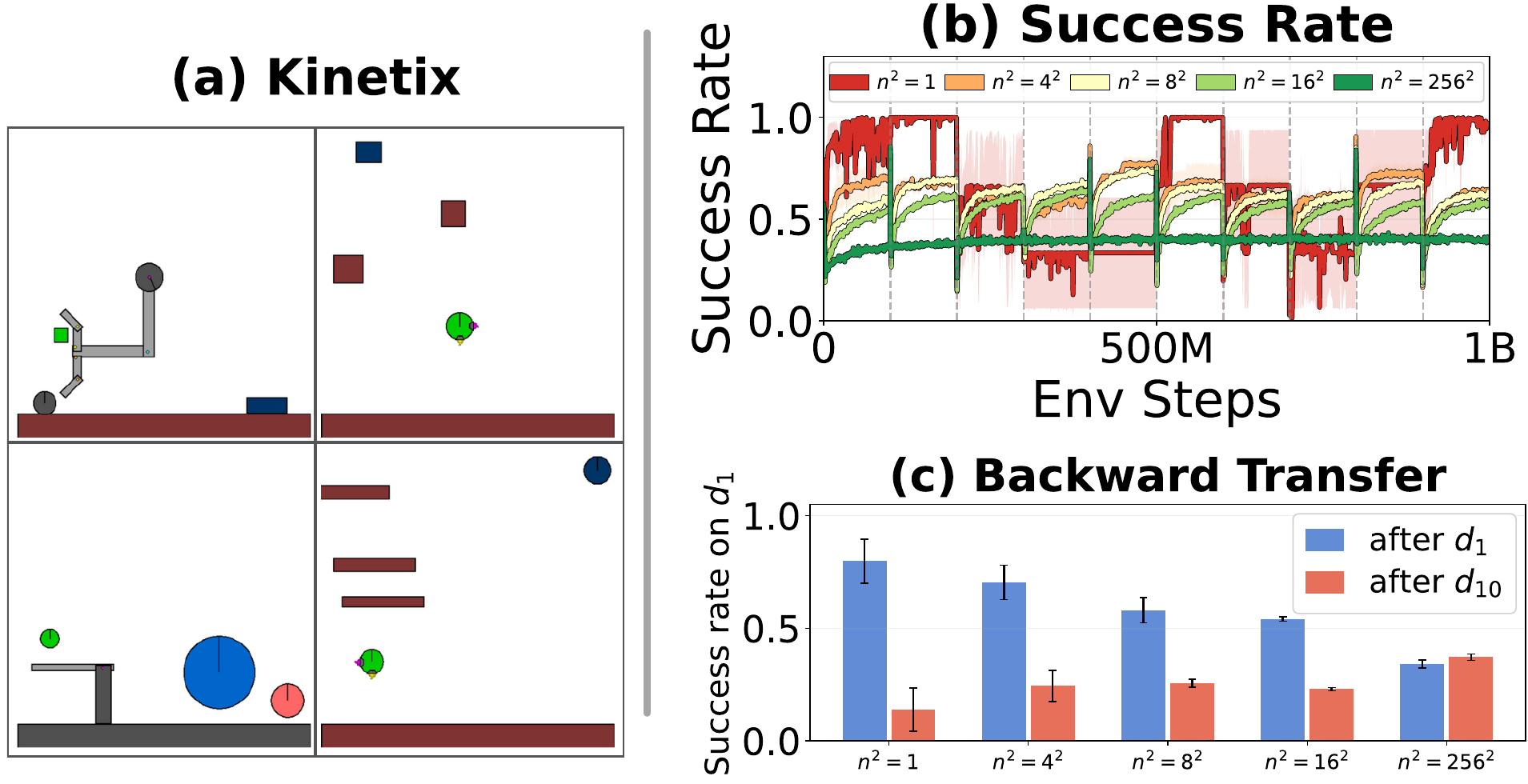}
    \caption{\revision{\textbf{Kinetix reproduces the same qualitative trade-off between diversity and continual RL observed in Banyan.} (a) Example Kinetix tasks. (b) Success rates across 9 distribution shifts as task diversity varies. (c) Backward transfer to the first task distribution after nine shifts.}}
    \label{fig:kinetix}
\end{figure}

\revision{\question{Do the effects of task diversity show up in a more realistic, less structured domain?}}
\revision{Banyan tasks are cleanly specified by task trees, so a natural question is whether the same phenomenon appears when they are not. We study Kinetix~\citep{kinetix}, an open-ended 2D physics-based control domain in which every task shares one goal: make the green and blue objects touch without the green object touching the red one. Kinetix procedurally generates millions of tasks, so we vary diversity by changing the number of random tasks in each task distribution from $n^2=1$ to $n^2=256^2$.
\textbf{Observations}. The agent receives a symbolic description of every shape, joint, and thruster in the scene.
\textbf{Actions}. The agent sets each motor to reverse, off, or forward and each thruster to off or on.
\textbf{Dynamics}. Rigid-body 2D physics with gravity, collisions, and the motors and thrusters the agent controls. \textbf{Reward}. The agent receives $+1$ when the green and blue objects touch and $-1$ when the green and red objects touch; either event ends the episode.
Kinetix is harder than Banyan, so success rates are lower throughout. The pattern, however, is the same. At intermediate diversity the agent keeps improving across task distributions, whereas at high diversity its improvement plateaus (\figref{fig:kinetix}, (b)). And just as in Banyan, task diversity improves backward transfer: $B(10,1)$ is strongly negative at $n^2=1$ and $n^2=16^2$ and turns positive at $n^2=256^2$ (\figref{fig:kinetix}, (c); \figref{fig:appendix-b-kinetix}). No forgetting is itself a gain, and success on $d_1$ after the full sequence is higher at $n^2=256^2$ than at $n^2=16^2$.}

\section{Discussion}

\revision{Our results present a paradox. Across a single distribution shift, increasing diversity along any axis drove the transfer gap toward zero---at the highest diversity, the agent entered each new distribution at the performance it had reached on the previous one. Across nine shifts the relationship shifted. The same diversity that smoothed every single boundary instead caused agents to plateau, and the highest diversity produced the least learning over the full sequence. This pattern persists with larger networks, across continual learning methods---i.e. continual backprop, specialized policy heads, replay---with Transformer memory, and in Kinetix---suggesting that it is not solely a problem of capacity, plasticity, shared-head interference, architecture, or of Banyan.}
\revision{Across these experiments, backward transfer reveals the same signature: performance on the first distribution keeps climbing while performance on each new distribution flatlines. This suggests that updates refine structure shared across distributions at the expense of fitting structure specific to each new distribution. What stalls is not learning, but specialization. We hypothesize that this may be a continual learning version of the ``primacy bias''~\citep{nikishin2022primacy}, where the policy an agent learns is biased by the features it first learned to represent. Here, the policy an agent learns as it encounters new task distributions seems ``primed''  by the task distributions it encountered early on. We leave studying this further to future work.}

\revision{\textbf{Limitations.} Our experiments span learning algorithms, memory architectures, and continual-learning methods, but we varied one component at a time. Combinations of these components could mitigate the negative effects of task diversity on continual RL, even though none individually resolved them. We also tested whether our findings extend to PointMass, a continuous-control domain, and Kinetix, an open-ended physics-based control domain, but all three domains remain far from real-world settings. Finally, although the effects persist with larger networks, we did not study architectures with trillions of parameters. It remains possible that \textit{more is different}~\citep{anderson1972more}, i.e. that sufficiently large architectures could enter a fundamentally different learning regime.}

\textbf{Conclusion.} \revision{\revision{More work is needed to determine whether task diversity inhibits continual RL in the real-world settings we care about. The current scaling paradigm in RL relies on relatively simple learning algorithms that benefit from increasingly diverse training data. Large RL models trained this way can generalize to new tasks, but they are typically deployed with frozen network parameters.} Our results suggest that the same diversity that \revision{supports generalization can also inhibit continual learning when parameters continue to change across a sequence of task distributions.} \revision{A key question is therefore} how to use diversity to promote continual reinforcement learning rather than inhibit it. \revision{We do not yet know whether our findings extend to real-world settings, but we hope Banyan supports the development of methods that address this challenge if they do.}} Learning and scale, on their own, may not be enough.

\section*{Acknowledgements}
We thank Natasha Jaques, Andrew Lampinen, and the Computational Cognitive Neuroscience group at Harvard for helpful discussions and feedback. We also thank the anonymous reviewers for their feedback. This work was supported by a gift from the Chan Zuckerberg Initiative Foundation to establish the Kempner Institute for the Study of Natural and Artificial Intelligence.

\bibliography{main}
\bibliographystyle{rlj}

\beginSupplementaryMaterials

\appendix

\section{Domains}
\label{sec:appendix-domains}

\subsection{Banyan}
\label{subsec:domains-banyan}

\textbf{World}. The Banyan world is a $G \times G$ grid of open floor tiles, walls, and objects. (We use $G = 6$.) The environment has up to 10 shapes and 12 colors, resulting in 120 shape--color pairs. Our experiment datasets use 8 shapes and 10 colors, for a total of 80 objects, and each task distribution draws its leaves from a pool of 40 of them. The agent holds at most one object at a time and begins each episode on a random open cell reachable from the top-left corner. There are nine discrete actions: the eight described earlier in the paper and a legacy action with no effect.

\textbf{Object interactions}. The environment has a fixed table that defines the outcome of object combinations and transformations. Combining objects, which can only be done via the rules of this table, does not require an action. When the inputs of a binary rule in the current task tree occupy adjacent cells, they are consumed and the output appears in the first input's cell. Newly produced objects can be combined again within the same step. Agents can also transform a single object by using the ``toggle'' action while standing on an adjacent tile or the same tile as the object. Since this combination and transformation table is fixed, the entire tree is determined by a topology and object assignment. To finish the task, an agent must pick up the root object after producing it. In a depth-1 task, the root is already present and only needs to be picked up.

\textbf{Dead-ends}. For each task distribution, each unordered pair of distinct objects in the distribution's leaf pool is marked as a distractor pair with probability $\rho$. (We use $\rho = 0.5$.) These pairs combine per the same fixed combination table mentioned above. The environment applies the current task's rules first, then checks distractor pairs around newly dropped objects. If a remaining pair involving the dropped object matches a distractor rule, the two merge and the episode ends with reward $-1$. These checks apply only to combinations, not transformations.

\textbf{Rewards}. The agent receives a success reward of $+1$. It also receives $-0.001$ per step and $+0.1$ the first time it picks up each object that appears in the task tree. Dead-end and unsuccessful timeout steps instead return only $-1$ and $0$, respectively.

\textbf{Observations}. Observations are flat vectors with three components. Every color one-hot has $13$ entries: the $12$ colors plus an ``empty'' value. For each grid cell: a one-hot over $20$ tile types, a color one-hot, and an indicator of whether the agent is on that cell. For the inventory: one bit per shape indicating whether it is held, plus a color one-hot for each shape that is zero when the shape is not held. For the goal: a one-hot over shape, a color one-hot, and a two-way one-hot indicating whether the task is pickup-only or requires producing an object. For $G = 6$, the single-agent observation has $1389$ dimensions: $1224$ for the grid, $140$ for the inventory, and $25$ for the goal. The agent does not observe the task tree, the rules, or a list of required intermediate objects. Intermediate objects are visible once produced.

\textbf{Layouts}. Layouts are wall masks over the grid. We sample a set of masks and retain those in which at least 30\% of cells are open, at least 55\% of open cells are reachable from the spawn region, and the cells around the spawn are clear. At $G = 6$, we sample between $2$ and $10$ wall cells. Masks are deduplicated exactly, and disjoint layout pools share no mask. At the start of an episode, a layout is drawn from the pool and the task tree's leaves are placed uniformly at random on reachable open cells.

\textbf{Reset}. Each reset samples a depth uniformly from $\{1, \dots, D\}$, then a task at that depth and a layout. \revision{We use $D = 6$ and a horizon of $256$ steps.} The task's rules are represented by fixed-shape tensors, so the step computation does not scan the task distribution. All environments are stepped in parallel with \texttt{jax.vmap}.

\textbf{Scale}. A pool of $P$ objects gives $\binom{P}{2}$ candidate depth-2 tasks, $3\binom{P}{4}$ candidate balanced depth-3 tasks, and $315\binom{P}{8}$ candidate balanced depth-4 tasks, using binary combinations and distinct leaves. These are combinatorial upper bounds before rejecting assignments with colliding object identities. For the full vocabulary of $P = 120$, the counts are roughly $2.5 \times 10^7$ at depth 3 and $2.6 \times 10^{14}$ at depth 4, before considering other topologies, transformations, or layouts. \revision{For the $P = 40$ pools used in our experiments, the counts are} $780$ at depth 2, $2.7 \times 10^5$ at depth 3, and $2.4 \times 10^{10}$ at depth 4.

\subsection{PointMass}
\label{subsec:domains-pointmass}

We constructed a continuous-control analog of Banyan to test whether our results generalize beyond grid worlds.

\textbf{World}. The agent is a point mass in a $[-1, 1]^2$ arena. In each step, it outputs a $2$-dimensional action vector $a \in [-1, 1]^2$ and moves by $0.12a$, clipped at the arena boundary. Eight fixed slots along the bottom of the arena hold objects drawn from a $120$-object leaf vocabulary. These include the task's required leaves and distractor objects in the remaining slots. Intermediate products have additional object identities. There is a single combine station near the top of the map and there are output slots above it. \revision{Trees have depth up to $2$.}

\textbf{Object interactions}. The agent picks up an object when it comes within $0.14$ of it and its inventory is empty. It holds at most one object at a time. It deposits the object when it comes within $0.14$ of an empty station. When the agent brings a second object to the station, the station merges the pair if the rulebook defines that combination. The rulebook is fixed and shared across tasks. A non-root output goes to an output slot, where the agent can pick it up and combine it again. If the rulebook does not define the pair, the first object remains in the station and the agent retains the second. The episode ends when the agent produces the root object, or picks it up in a depth-1 task.

\textbf{Dead-ends}. The episode also ends if the agent performs a merge that the rulebook allows but the current tree does not require.

\textbf{Rewards}. The agent receives $+1$ for completing the task and $-1$ for a dead-end. It loses $0.005$ each time it attempts a merge the rulebook does not define, $0.001$ per step, and $0.001\|a\|^2$ for control effort.

\textbf{Observations}. The agent observes the positions and contents of the leaf slots, output slots, and station, its own position and velocity, the object it holds, the goal object, and the elapsed fraction of the episode. It does not observe the tree or the rulebook.

\textbf{Training}. We train PPO with a $256$-unit tanh MLP over a per-object encoder and a tanh-squashed Gaussian policy. We use learning rate $3 \times 10^{-4}$, $\gamma = 0.99$, $\lambda = 0.95$, clip $0.2$, entropy coefficient $10^{-3}$, value coefficient $0.5$, gradient-norm clipping at $0.5$, $8$ minibatches, and $4$ epochs. We run $1024$ environments with $64$-step rollouts and a $64$-step episode horizon \revision{for $2 \times 10^{8}$ steps in total across the two distributions}. We construct $d_1$ from $n$ full-tree object assignments and $d_2$ from $1000$, including shallower subtasks from each assignment.

\subsection{Kinetix}
\label{subsec:domains-kinetix}

Kinetix~\citep{kinetix} is a GPU-vectorized 2D rigid-body physics environment; see the original paper for the observation and action spaces and the level format. We use the ``small'' preset, which has $5$ polygons, $2$ circles, $1$ joint, and $1$ thruster. Every task has the same goal: the green shape must touch the blue shape without touching the red shape. The episode ends with $+1$ when green touches blue and $-1$ when green touches red. Otherwise, the agent receives a dense shaping reward based on green--blue distance (scale $0.2$). Episodes last at most $256$ steps.

\textbf{Task sampling}. We sample levels from Kinetix's default procedural generator and exclude those a no-op policy solves within $64$ steps. \revision{Each task distribution is a seeded pool of $n^2$ levels, so $n^2$ controls diversity exactly as in Banyan}, and the pools are disjoint across the ten distributions in the sequence. Because Kinetix folds the training seed into the level seed, different training seeds see different level pools.

\textbf{Training}. We train a single transformer policy with PPO across $10$ sequential distributions for $10^{8}$ steps per distribution. \revision{We sweep $n^2 \in \{1, 4^2, 8^2, 16^2, 256^2\}$ and run $3$ seeds for each.}

\section{Algorithms}
\label{sec:appendix-algorithms}

This section defines the reinforcement learning algorithms, memory architectures, and continual learning methods we evaluate. Network sizes, hyperparameters, and search procedures are in Section~\ref{sec:appendix-implementation}.

\subsection{Problem setting}
\label{subsec:problem-setting}

We formulate each Banyan task $\task$ as a Partially Observable Markov Decision Process~\citep[POMDP;][]{kaelbling1998planning} $\langle \mathcal{S}, \mathcal{A}, \mathcal{X}, P, O, R\rangle$, where $\mathcal{S}$ denotes the environment state space, $\mathcal{A}$ denotes the action space, and $\mathcal{X}$ denotes (potentially partial) observations of the environment. When the agent takes an action $a_t \in \mathcal{A}$ in state $s_t \in \mathcal{S}$, the next state is sampled according to $s_{t+1} \sim P(\cdot \mid s_t, a_t)$, an observation is generated via $x_{t+1} = O(s_{t+1})$, and the agent receives reward $r_{t+1} = R(s_{t+1})$. All tasks share the same state space, action space, observation function, and dynamics $P$ (which objects combine to produce what); a task's objects and topology (\S\ref{sec:domain}) determine its reward function $R$, and its layout determines the initial state. The observation includes the task's goal item, so a single policy can act across tasks.

The agent never observes $s_t$ directly. Instead, it maintains a learned representation of its history, $h_t = f_\theta(x_{\le t}, a_{<t})$, computed from the observations and actions so far by a memory architecture; here $f_\theta$ is either a recurrent network or a transformer (Section~\ref{subsec:memory-architectures}). Throughout, $\theta$ denotes all learnable parameters, including those of $f_\theta$. The agent selects actions with a policy $\pi_\theta(a \mid h_t)$. We call the discounted sum of future rewards the \emph{return}, $G_t = \sum_{k \ge 0} \gamma^k r_{t+k+1}$, with discount factor $\gamma \in [0,1)$. The agent's objective is a policy that maximizes expected return.

A task distribution $d$ is a distribution over tasks (\S\ref{sec:experiments}). The agent faces a sequence of task distributions $d_1, \dots, d_K$. During phase $i$, the task for each episode is sampled as $\task \sim d_i$, and the agent trains for $N$ environment steps before the distribution changes to $d_{i+1}$. Parameters and optimizer state carry over across phases; nothing is reset at a shift. We study agents that continue learning across task distributions, i.e. that keep updating their parameters $\theta$.

\subsection{Reinforcement learning algorithms}
\label{subsec:rl-algorithms}

Both algorithms we consider learn from rollouts. The agent acts in many parallel environments with its current policy for $T$ steps, and the memory architecture processes each resulting trajectory $(x_1, a_1, r_2, \ldots, x_T)$ into history representations $(h_1, \ldots, h_T)$. Each algorithm then constructs a learning target from the rewards observed within the rollout, combined with its own predictions about what lies beyond it, and takes several gradient passes over the rollout before collecting another. Throughout, $d_t$ is $1$ if the transition after taking $a_t$ ends the episode and $0$ otherwise. The two algorithms differ in what they learn: PPO learns a policy directly and uses a value function only to judge which actions to reinforce, whereas PQN learns an action-value function and derives its policy from it.

\subsubsection{Policy-based: PPO}
\label{subsubsec:ppo}

Proximal Policy Optimization~\citep[PPO;][]{schulman2017proximal} learns a policy $\pi_\theta(a \mid h_t)$, the actor, together with a value function $V_\theta(h_t)$, the critic, which approximates the expected return under the current policy, $V^\pi(h_t) = \mathbb{E}_\pi[G_t \mid h_t]$. The critic's role is to judge the actor. An action is reinforced in proportion to its \emph{advantage}, $A^\pi(h_t, a_t) = \mathbb{E}_\pi[G_t \mid h_t, a_t] - V^\pi(h_t)$: how much better the return after taking $a_t$ is than the critic expected from $h_t$.

The advantage must be estimated from a finite rollout. The simplest estimate is the one-step temporal-difference (TD) error,
\begin{align}
    \delta_t = r_{t+1} + \gamma(1-d_t)\, V_\theta(h_{t+1}) - V_\theta(h_t),
    \label{eq:td-error}
\end{align}
which compares one observed reward plus the critic's estimate of everything after it against the critic's estimate at $h_t$. This estimate is ``bootstrapped'': it leans on the very function being learned, so it inherits the critic's bias but has low variance. Summing observed rewards to the end of the rollout before bootstrapping has the opposite character. Generalized advantage estimation~\citep[GAE;][]{schulman2016gae} interpolates between the two with an exponentially weighted sum of TD errors,
\begin{align}
    \hat{A}_t = \sum_{l=0}^{T-t-1} (\gamma\lambda)^l\, \delta_{t+l},
    \label{eq:gae}
\end{align}
where $\lambda \in [0,1]$ controls the weighting: $\lambda = 0$ recovers \eqref{eq:td-error}, and $\lambda = 1$ recovers the full-rollout estimate that bootstraps only from $V_\theta(h_T)$.

Given advantages, PPO updates the actor over several passes through the same rollout. Because the policy changes between passes, the data was collected by an older policy $\pi_{\mathrm{old}}$. PPO accounts for this with the probability ratio $\rho_t(\theta) = \pi_\theta(a_t \mid h_t) / \pi_{\mathrm{old}}(a_t \mid h_t)$, and it prevents any single rollout from moving the policy too far by clipping that ratio:
\begin{align}
    L_{\tt policy}(\theta) = -\frac{1}{T}\sum_{t=1}^{T} \min\Big(
        \rho_t(\theta)\hat{A}_t,\;
        \operatorname{clip}\!\big(\rho_t(\theta), 1-\epsilon_{\tt clip}, 1+\epsilon_{\tt clip}\big)\hat{A}_t
    \Big).
    \label{eq:ppo-policy}
\end{align}
The critic is regressed onto the return implied by the same advantage estimate,
\begin{align}
    L_{\tt value}(\theta) = \frac{1}{T}\sum_{t=1}^{T} \Big(
        V_\theta(h_t) - \operatorname{stopgrad}\big(\hat{A}_t + V_\theta(h_t)\big)
    \Big)^2,
    \label{eq:ppo-value}
\end{align}
where $\operatorname{stopgrad}$ blocks gradients through its argument.\footnote{As is common in PPO codebases, we also clip the change in $V_\theta$ between passes, an empirical trick that limits how far the critic moves on a single rollout.} Finally, an entropy bonus $\mathcal{H}[\pi_\theta(\cdot \mid h_t)]$ discourages the policy from collapsing onto a few actions before it has explored. The complete loss is $L_{\tt policy} + c_v L_{\tt value} - c_e \mathcal{H}$, with coefficients given in Section~\ref{sec:appendix-implementation}.

\subsubsection{Value-based: PQN}
\label{subsubsec:pqn}

Parallelised Q-Network~\citep[PQN;][]{Gallici25simplifying} learns an action-value function $Q_\theta(h_t, a)$ that approximates the optimal action-value function, $Q^*(h_t, a) = \max_\pi \mathbb{E}_\pi[G_t \mid h_t, a_t = a]$: the expected return from taking action $a$ at $h_t$ and acting optimally thereafter. The policy is derived from it. The agent selects the action with the highest predicted value, except with probability $\epsilon_{\tt explore}$, when it selects a random action.

As in Q-learning~\citep{sutton2018}, $Q_\theta$ is trained to match a bootstrapped target. The 1-step target, $r_{t+1} + \gamma(1-d_t)\max_a Q_\theta(h_{t+1}, a)$, combines one observed reward with the network's own estimate of the best return thereafter. As with GAE, PQN spreads the weight across longer stretches of observed reward using the $Q(\lambda)$ target~\citep{peng1994incremental}, computed backward through the rollout from a final bootstrap:
\begin{align}
    y^\lambda_T &= \max_a Q_\theta(h_T, a), \nonumber \\
    y^\lambda_t &= r_{t+1} + \gamma(1-d_t)\Big[(1-\lambda)\max_a Q_\theta(h_{t+1}, a) + \lambda\, y^\lambda_{t+1}\Big].
    \label{eq:qlambda}
\end{align}
Here $\lambda$ plays the same role as in GAE: $\lambda = 0$ recovers the 1-step target and $\lambda = 1$ sums the observed rewards to the rollout's end before bootstrapping. The network minimizes the squared error to this target,
\begin{align}
    L_{\tt Q}(\theta) = \frac{1}{T}\sum_{t=1}^{T} \Big(
        Q_\theta(h_t, a_t) - \operatorname{stopgrad}\big(y^\lambda_t\big)
    \Big)^2,
    \label{eq:pqn-loss}
\end{align}
where the same network computes both prediction and target but gradients pass only through the prediction. PQN uses no target network or replay buffer, relying instead on layer normalization~\citep{ba2016layer} inside the network and on fresh rollouts from many parallel environments.

\subsubsection{Memory architectures}
\label{subsec:memory-architectures}

Neither algorithm specifies how the history $h_t$ is represented. Banyan tasks are partially observable, so the agent must carry information across steps: which objects it has already combined, what it holds, and what it has seen. A memory architecture maps the sequence of observations up to time $t$ to a fixed-size state that the actor, critic, or Q head reads. We evaluate two families.

\textbf{Recurrent memory}. A recurrent network updates a hidden state one step at a time from the previous state, the current observation, and the previous action, so the state at time $t$ summarizes everything before it. Because the update is sequential, the cost per step is constant, but information from early in an episode must survive every intermediate update to remain available. We use a gated recurrent unit~\citep[GRU;][]{cho2014gru} for PPO and a long short-term memory network~\citep[LSTM;][]{hochreiter1997lstm} for PQN. The state is reset at episode boundaries.

\textbf{Attention-based memory}. An attention-based memory~\citep{parisotto2020stabilizing}, here a Transformer-XL, instead keeps the representations of a fixed window of past steps and attends over them directly, so information from any step in the window is available without passing through intermediate states. Relative positional encoding lets the same attention weights apply at every offset within the window, and the cached representations of earlier segments extend the effective context beyond a single rollout. Attention does not cross episode boundaries.

\subsection{Continual learning algorithms}
\label{subsec:cl-algorithms}

Each method below modifies the PPO agent to reduce a specific failure of continual learning. Continual backpropagation changes how the network is maintained during training, replay changes which data the network trains on, and multi-head policies change which parameters each distribution updates. The PPO objective is otherwise unchanged.

\textbf{Continual backprop}. Continual backpropagation~\citep[CBP;][]{dohare2024loss} modifies gradient-based training to address loss of plasticity (a trained network's reduced ability to learn from new data). It continually replaces a small fraction of low-utility hidden units rather than reinitializing the whole network. We estimate each unit's utility from how much its activation varies across the minibatch and the magnitude of its outgoing weights:
\[
u_i \leftarrow \eta u_i + (1-\eta)\,a_i w_i,
\]
where $a_i$ is the mean absolute deviation of unit $i$'s activation from its minibatch mean, $w_i$ is the sum of the mean absolute outgoing weights over its destination weight matrices, and $\eta$ controls the averaging timescale. This gives low utility to network units whose activations vary little or whose outputs have little influence on downstream computations.

A unit becomes eligible for replacement only after a maturity threshold, giving new units time to learn before they are assessed. After each gradient step, CBP selects the lowest-utility eligible units for replacement. It samples new incoming weights, zeros their incoming biases and outgoing weights, and resets their ages, utility estimates, and associated optimizer moments. It also adjusts downstream biases where available to compensate for the removed units' estimated mean contribution. Zeroing the outgoing weights ensures that a newly generated unit does not immediately change the network's output, while the new incoming weights introduce a fresh feature that can subsequently be learned. The remaining network continues learning with the same PPO objective.

\textbf{Replay}. Replay addresses forgetting by ensuring that earlier distributions continue to contribute gradients after the agent has moved on to later ones. Once the agent finishes its training budget on a distribution, it continues training for a further, smaller budget on an earlier distribution, so no distribution is ever the sole source of updates for long. Generative replay~\citep{shin2017generative} trains a model of past data and samples synthetic experience from it. We use generative replay with a perfect model of the environment: the agent revisits the earlier distribution itself and collects new on-policy experience there with its current policy. Fresh experience means the agent trains on the distribution rather than on a fixed set of recorded episodes, so what it retains is the distribution's structure rather than the specifics of any recording. The objective is the same clipped PPO loss.

\textbf{Multi-head}. Multi-head policies~\citep{kessler2022same, caccia2023taskagnostic} address interference at the output by giving each distribution its own action-selection parameters while sharing everything upstream. The embedding, memory, actor hidden layer, and critic form a shared body that receives gradients from every distribution. The final layer that maps the actor's hidden representation to action logits is duplicated, one copy per distribution, and the copy matching the current distribution is selected at each forward pass. Learning a new distribution cannot overwrite an earlier distribution's action mapping, but it can still change the shared representation that mapping reads. The method requires the distribution identity at training and evaluation time.

\section{Implementation details}
\label{sec:appendix-implementation}

\subsection{PPO-based algorithms}
\label{subsec:ppo-implementation}

All PPO agents on Banyan share one architecture and one update rule; the only differences between variants are the memory module and, for the continual learning methods, the modifications described in Section~\ref{subsec:cl-implementation}. The network is a $1024$-unit ReLU observation embedding, a linear projection to $512$ units, a $512$-unit memory module, and separate $2048$-unit ReLU actor and critic heads. GAE uses $\gamma = 0.99$ and $\lambda = 0.8$, with per-minibatch advantage normalization. Each rollout of $256$ steps from $1024$ parallel environments is trained for $4$ epochs over $4$ trajectory minibatches with policy clip $\epsilon_{\mathrm{clip}} = 0.2$, value clip $10$, value coefficient $0.8$, \revision{and entropy coefficient $0.0333$. We use AdamW (weight decay $10^{-4}$) at a constant learning rate of $1.67 \times 10^{-4}$ and clip the global gradient norm at $0.25$.} Memory is reset at episode boundaries.

\textbf{RNN-based memory}. The memory module is a $512$-unit GRU. Its hidden state is carried across rollout boundaries within an episode and reset to zero when an episode ends.

\textbf{Transformer-based memory}. We replace the GRU with a Transformer-XL--style network with $2$ layers, $8$ heads, model and QKV width $512$, relative positional attention, and memory for the previous $128$ steps. Attention includes the current step and does not cross episode boundaries. We do not use GTrXL gating~\citep{parisotto2020stabilizing}. \revision{The embedding, actor and critic heads, and PPO update settings are the same as for the GRU agent, except that the transformer runs use the PPO + Replay configuration (Adam at $1.44 \times 10^{-4}$ with a $5\%$ linear learning-rate warmup and entropy coefficient $0.0303$).} \revision{Our implementation is adapted from an open-source JAX implementation of TrXL-PPO.}\footnote{\url{https://github.com/Reytuag/transformerXL_PPO_JAX}}

\subsection{PQN}
\label{subsec:pqn-implementation}

The network applies batch renormalization to the input, followed by a $512$-unit layer-normalized ReLU layer. It concatenates this representation with a one-hot of the previous action and passes the result through a $512$-unit LSTM and a linear Q head. We use $\gamma = 0.99$, $\lambda = 0.535$, and $4$ epochs over $4$ minibatches per $256$-step rollout from $512$ environments. Exploration is $\epsilon$-greedy, with $\epsilon_{\mathrm{explore}}$ linearly annealed from $1$ to $0.087$. The exploration counter is carried across distributions rather than restarted at each boundary. We use RAdam at $1.09 \times 10^{-4}$ and clip the global gradient norm at $0.25$.

\subsection{Continual learning algorithms}

\label{subsec:cl-implementation}

\revision{All three methods build on the PPO + RNN architecture above. Each paragraph states what the method changes; Section~\ref{subsec:hyperparameter-search} describes how its settings were chosen.}

\textbf{Continual backprop}. We apply CBP to the embedding, GRU, actor hidden layer, and critic hidden layer. \revision{We use utility decay $\eta = 0.986$, maturity $5000$ gradient steps, and a replacement rate of $5 \times 10^{-5}$ of eligible units per gradient step, with fractional counts accumulated across steps. The hyperparameter search (Section~\ref{subsec:hyperparameter-search}) selected AdamW ($\beta_1 = 0.9$, $\beta_2 = 0.999$, weight decay $10^{-4}$) at $2.0 \times 10^{-4}$ with entropy coefficient $0.0333$ and the same network widths as PPO + RNN ($1024$-unit embedding, $512$-unit GRU, and $2048$-unit heads). The remaining PPO settings are unchanged.}

\textbf{Replay}. After the on-policy training block on each distribution $d_k$ with $k \geq 2$, the agent trains for an additional $0.2 \cdot N$ steps, where $N$ is the per-distribution budget, on one earlier distribution drawn uniformly from $\{d_{1}, \ldots, d_{k-1}\}$. The agent collects this block's experience by running its current policy in that distribution's environment; nothing is stored between blocks. Replay therefore adds $20\%$ to the total environment steps. The architecture and PPO update settings are identical to PPO + RNN. The optimizer is Adam at a constant learning rate of $1.44 \times 10^{-4}$ with entropy coefficient $0.0303$.

\textbf{Multi-head}. The actor output layer has one $|\mathcal{A}|$-unit block per distribution, ten in total for the ten-distribution sequence, and the block indexed by the current distribution produces the action logits at each forward pass. The embedding, GRU, actor hidden layer, and critic are shared across all distributions and receive gradients from every one. The distribution index is provided at training and evaluation time. The network widths, optimizer, and all PPO settings are identical to PPO + Replay.

\subsection{Hyperparameter search}

\label{subsec:hyperparameter-search}

Every random search below used the same protocol. Each search drew $15$ configurations, and each configuration sampled every searched axis independently from its range. We sampled learning rates and the CBP replacement rate log-uniformly, the maturity threshold uniformly over integers, and every other axis uniformly over its interval. We evaluated each configuration with $3$ seeds by success on the final distribution and held all other settings fixed. The search setup used three distributions with tasks at depths $1$ through $6$, including $128$ depth-$6$ tasks per distribution, from the full $120$-object vocabulary. The depth-$6$ task topologies and producer signatures were disjoint across distributions. Each distribution received $10^{8}$ training steps.

\textbf{PPO + RNN}. We searched the learning rate over $[10^{-5}, 3 \times 10^{-4}]$, the entropy coefficient over $[0, 0.1]$, and the network size, with embedding dimension in $\{512, 1024\}$, memory dimension in $\{256, 512\}$, and actor and critic hidden dimension in $\{1024, 2048\}$. The search and the final configuration both use AdamW with weight decay $10^{-4}$. The search selected the larger network.

\textbf{PPO + RNN + CBP}. CBP starts from the PPO + RNN configuration. We searched the learning rate over $[10^{-5}, 10^{-3}]$ and CBP's own settings: the utility decay over $[0.95, 0.999]$, the maturity threshold over $[100, 5000]$, and the replacement rate over $[4 \times 10^{-6}, 3 \times 10^{-2}]$.

\textbf{PPO + Replay}. We searched the learning rate over $[10^{-5}, 3 \times 10^{-4}]$ and the entropy coefficient over $[0, 0.1]$, with the revisit budget ($0.2$) fixed and the revisited distribution drawn uniformly from all earlier ones. The search used AdamW with weight decay $10^{-4}$; the final configuration uses Adam.

\textbf{PPO + Multi-head}. Multi-head uses the PPO + Replay configuration.

\textbf{PPO + Transformer}. The Transformer starts from the PPO + Replay configuration. We screened $14$ configurations on a two-distribution sequence at $n = 256$ with one seed. We varied width ($256$ or $512$), depth ($2$ or $4$ layers), attention window ($128$ or $255$ previous steps), and GTrXL gating (on or off). The screen also included a configuration with a lower learning rate ($5 \times 10^{-5}$). Disabling gating improved first-distribution success from $0.64$ to $0.76$, and width $512$ outperformed $256$. We selected no gating, width $512$, $2$ layers, and a $128$-step memory window.

\textbf{PQN}. We searched the learning rate over $[5 \times 10^{-5}, 10^{-3}]$, $\lambda$ over $[0.1, 0.95]$, and the final $\epsilon_{\mathrm{explore}}$ over $[0.01, 0.2]$.

\section{Experiment Details}
\label{sec:appendix-experiments}

\subsection{Two-distribution experiments}
\label{subsec:two-distribution-experiments}

\textbf{Experiment construction}. We train PPO and PQN on distributions $d_1$ and then $d_2$, with $N = 10^8$ environment steps per distribution. We sweep the number of layouts, object assignments, and task-tree topologies in $d_1$, one axis at a time. For layouts and object assignments, we use $n \in \{1, 10, 100, 1000, 10000\}$. For topologies, we use $n \in \{1, 2, 4, 8, 16, 32, 64, 128, 256\}$. We keep the size of $d_2$ fixed as we vary $n$ and use held-out examples along the tested axis. Changing $n$ changes the training diversity, not the training budget.

\subsection{Ten-distribution continual learning}
\label{subsec:ten-distribution-experiments}

\textbf{Task sequence construction}. We train on
$d_1, \dots, d_{10}$ with $N = 10^8$ environment steps per task distribution,
for $10^9$ steps in total. Each distribution contains $n$ depth-$6$ task
trees and $n$ layouts, giving $n^2$ depth-$6$ task--layout combinations.
Each tree also contributes one nested subtask at each depth from $1$
through $5$, extracted along a longest root-to-leaf path. We count
diversity using the depth-$6$ tasks. Layout pools and depth-$6$ topologies
are disjoint across distributions. We also exclude repeated transformations
and input combinations across distributions, while allowing individual
object identities to recur. We use
$n \in \{1, 2, 4, 8, 16, 32, 64, 128, 256\}$ \revision{for Figure~\ref{fig:ten-dist-top} and $n \in \{1, 16, 256\}$ for Figures~\ref{fig:methods} and \ref{fig:transformer}}.

\textbf{Training protocol}. We train on each distribution for $N$ steps and then, at each shift, change the task distribution and continue training without resetting the network parameters or optimizer state. For CBP, we also retain unit ages and utility estimates. We keep the algorithm configuration fixed across distributions within a run.

\section{Additional results}
\label{sec:appendix-additional-results}

\subsection{Transfer gap}
\begin{figure}[h]
    \centering
    \includegraphics[width=0.9\linewidth]{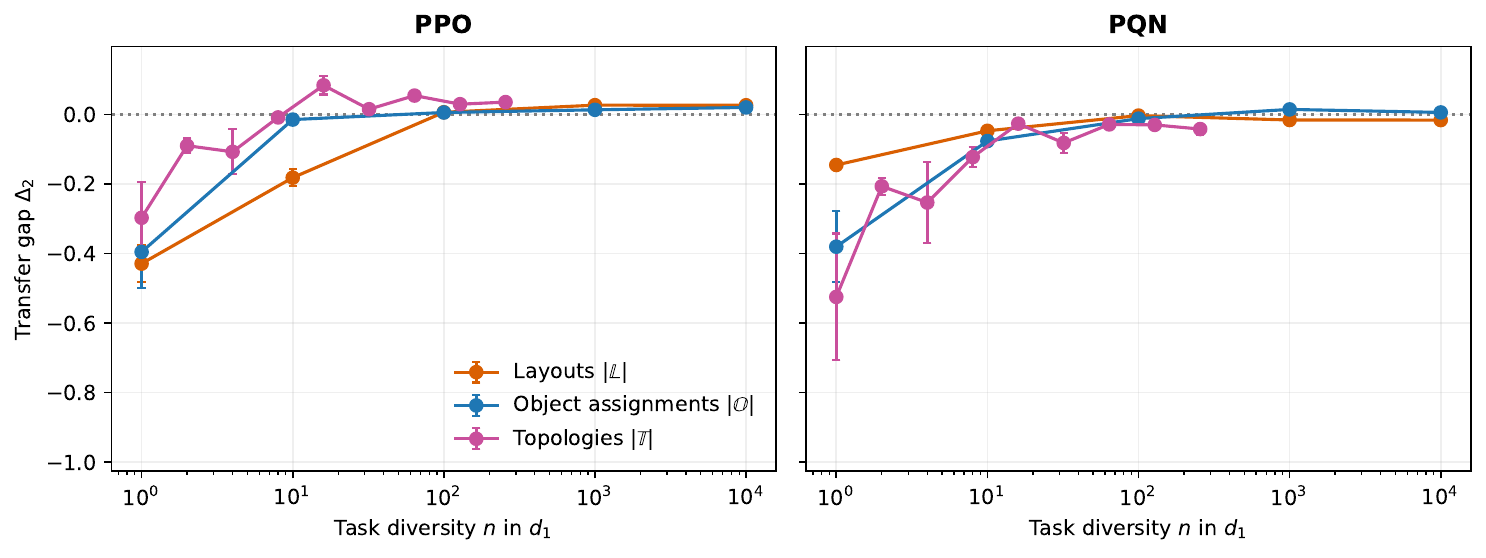}
    \caption{\revision{\textbf{Increasing diversity in $d_1$ drives the transfer gap toward zero along every axis.} Transfer gap $\Delta_2 = S_{\text{start}}(d_2) - S_{\text{end}}(d_1)$ as the number of layouts, object assignments, or topologies in $d_1$ grows, for PPO (left) and PQN (right), averaged over depths 1--6. Error bars are standard error of the mean across seeds.}}
    \label{fig:appendix-transfer-gap}
\end{figure}

\revision{\figref{fig:appendix-transfer-gap} plots the transfer gap $\Delta_2$ behind Figure~\ref{fig:r1} directly, averaged over depths 1--6.}

\clearpage
\subsection{Backward transfer values}
\begin{figure}[h]
    \centering
    \includegraphics[width=\linewidth]{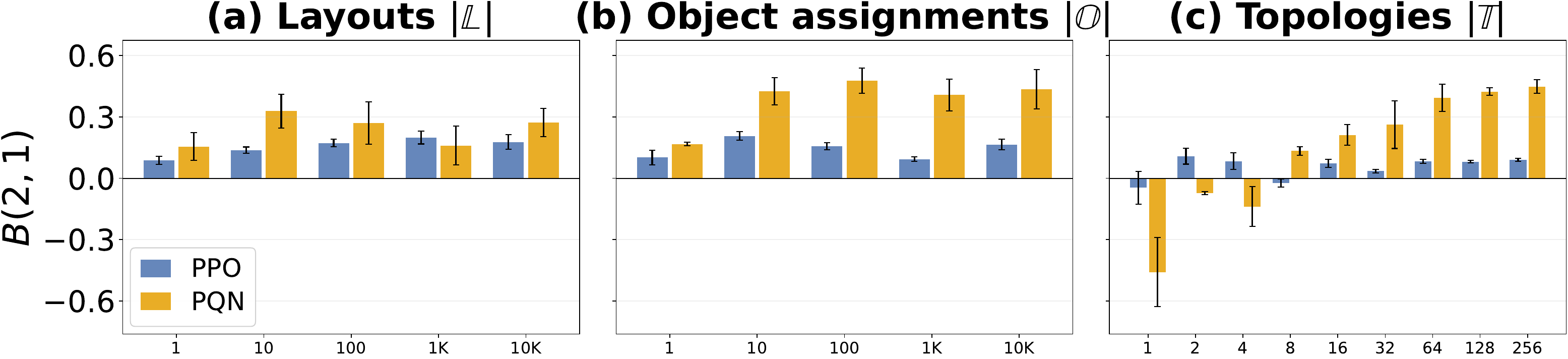}
    \caption{\revision{\textbf{Backward transfer in the two-distribution experiment.} $B(2,1)$ as the number of layouts, object assignments, or topologies in $d_1$ grows, for PPO and PQN; companion to Figure~\ref{fig:r2}.}}
    \label{fig:appendix-b-independent}
\end{figure}

\begin{figure}[h]
    \centering
    \includegraphics[width=0.45\linewidth]{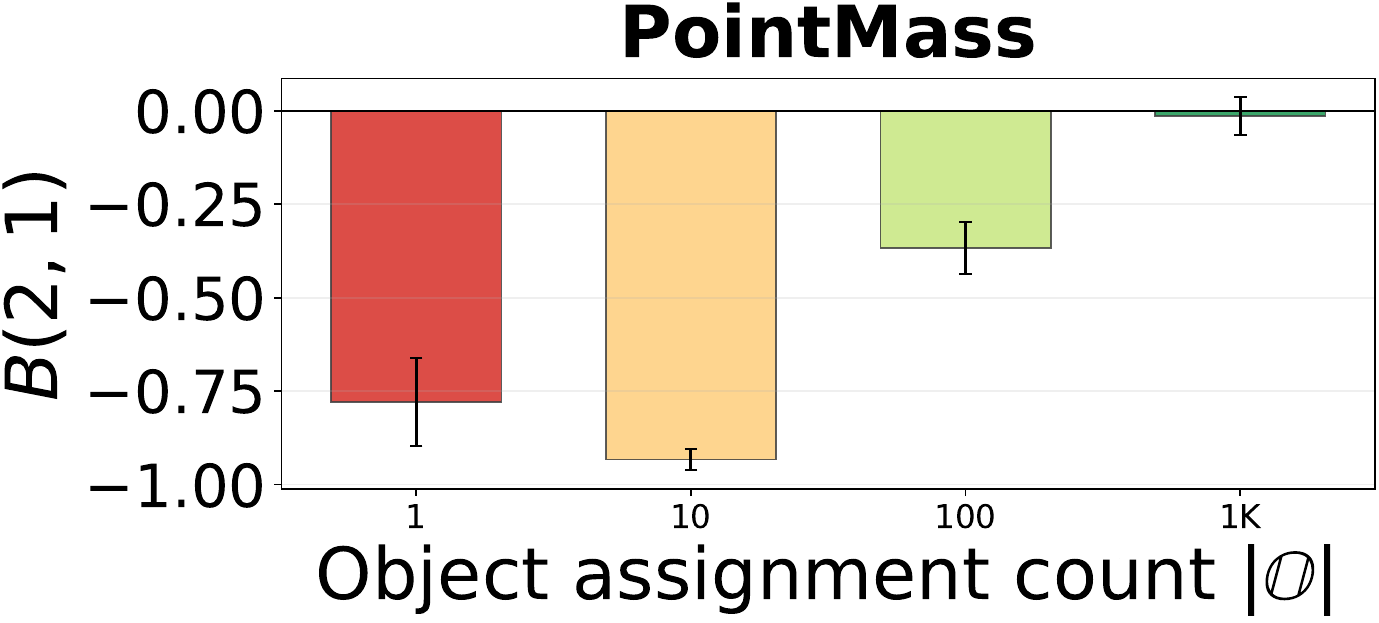}
    \caption{\revision{\textbf{Backward transfer in PointMass.} $B(2,1)$ per object-assignment count; companion to Figure~\ref{fig:continuous-banyan}.}}
    \label{fig:appendix-b-pointmass}
\end{figure}

\begin{figure}[h]
    \centering
    \includegraphics[width=0.53\linewidth]{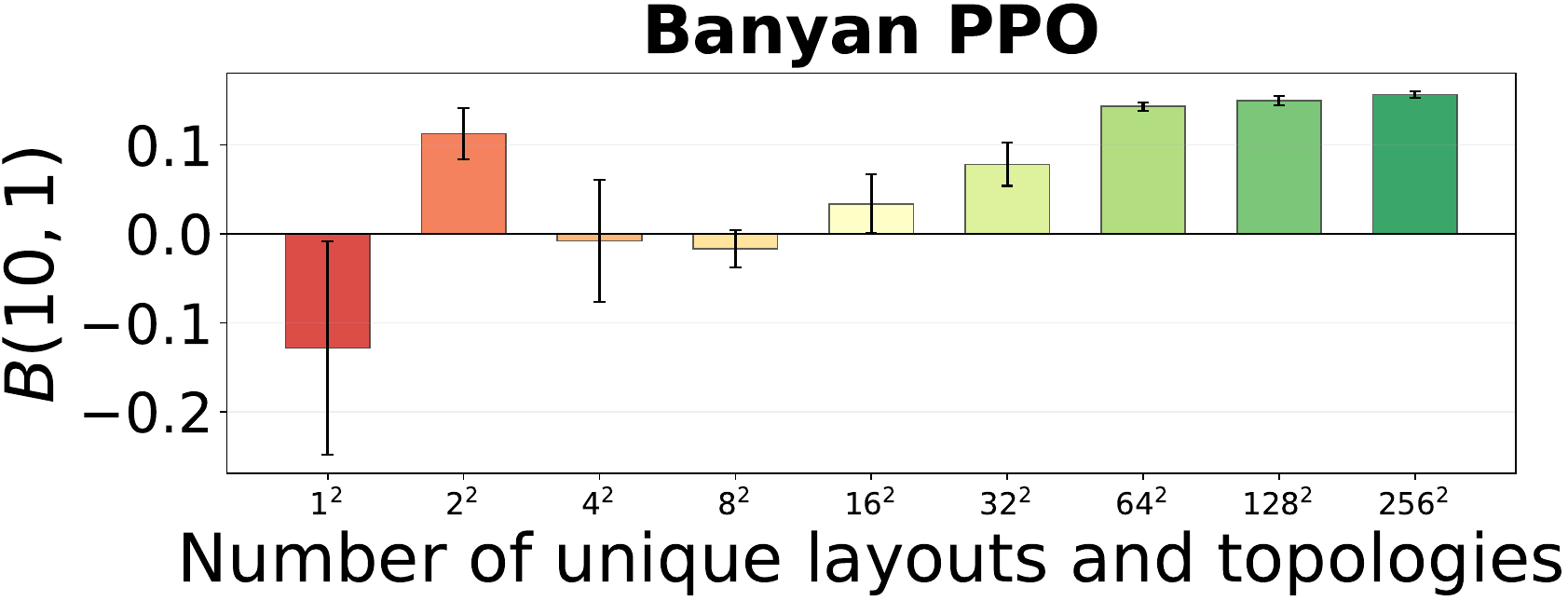}
    \caption{\revision{\textbf{Backward transfer for PPO across ten task distributions.} $B(10,1)$ per diversity level; companion to Figure~\ref{fig:ten-dist-top}.}}
    \label{fig:appendix-b-ppo}
\end{figure}

\begin{figure}[h]
    \centering
    \includegraphics[width=\linewidth]{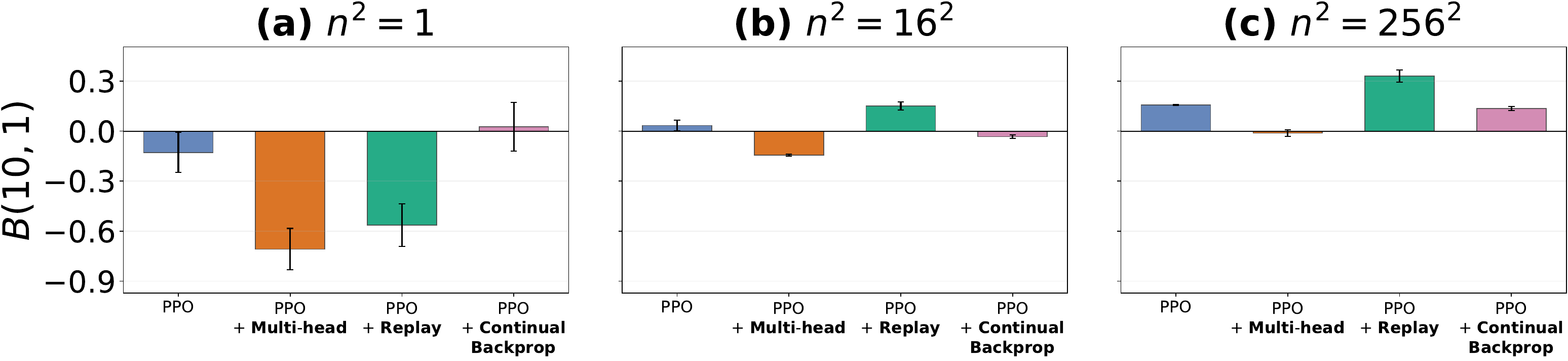}
    \caption{\revision{\textbf{Backward transfer for the continual-learning methods.} $B(10,1)$ per method at each diversity level; companion to Figure~\ref{fig:methods}.}}
    \label{fig:appendix-b-methods}
\end{figure}

\begin{figure}[h]
    \centering
    \includegraphics[width=\linewidth]{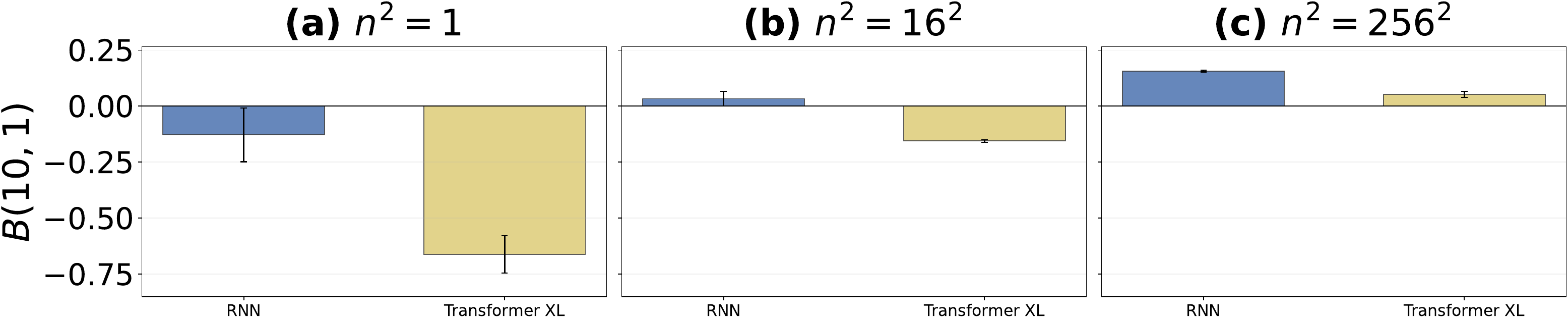}
    \caption{\revision{\textbf{Backward transfer for RNN versus Transformer-XL memory.} $B(10,1)$ at each diversity level; companion to Figure~\ref{fig:transformer}.}}
    \label{fig:appendix-b-transformer}
\end{figure}

\begin{figure}[h]
    \centering
    \includegraphics[width=0.6\linewidth]{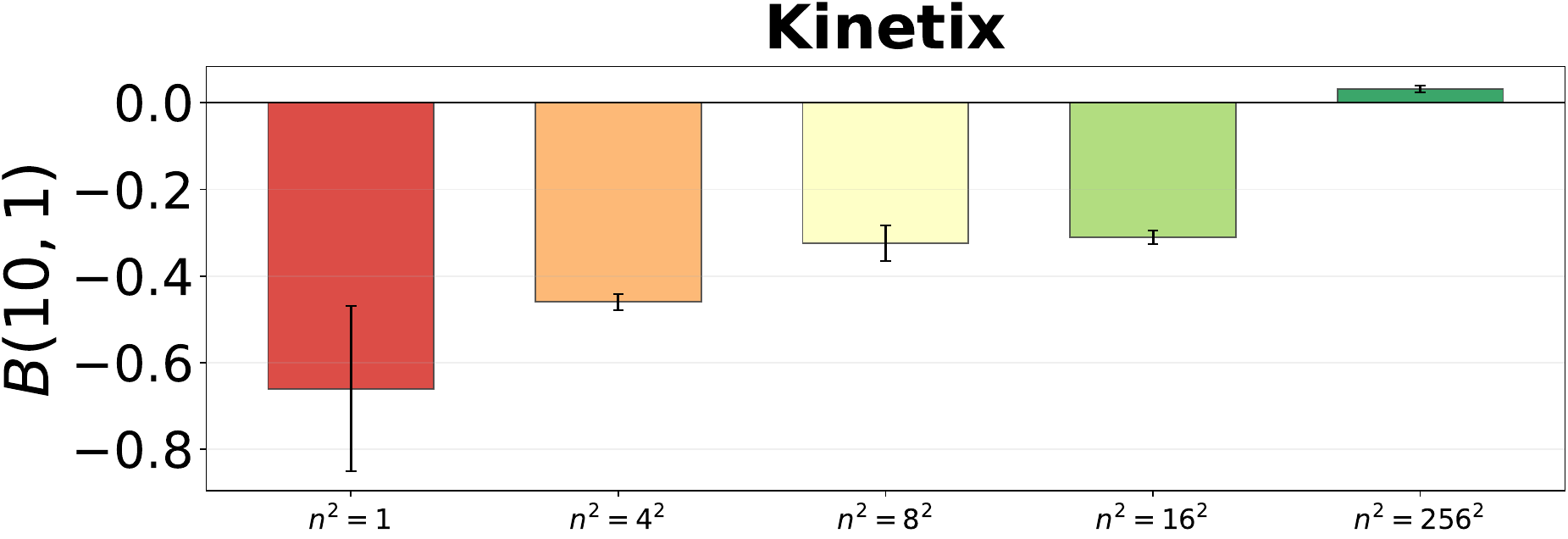}
    \caption{\revision{\textbf{Backward transfer in Kinetix.} $B(10,1)$ per task-pool size; companion to Figure~\ref{fig:kinetix}.}}
    \label{fig:appendix-b-kinetix}
\end{figure}

\revision{Figures~\ref{fig:r2}, \ref{fig:continuous-banyan}, \ref{fig:ten-dist-top}, \ref{fig:methods}, \ref{fig:transformer}, and \ref{fig:kinetix} report backward transfer as two success rates on $d_1$. Figures~\ref{fig:appendix-b-independent}--\ref{fig:appendix-b-kinetix} show the corresponding signed values $B(k,1) = S_{\text{end}}(d_1; k) - S_{\text{end}}(d_1; 1)$, where $k$ is the final distribution, so positive values mean later training improved success on $d_1$. Bars are means across seeds; error bars are standard error across seeds.}

\subsection{Network capacity}
\begin{figure}[h]
    \centering
    \includegraphics[width=\linewidth]{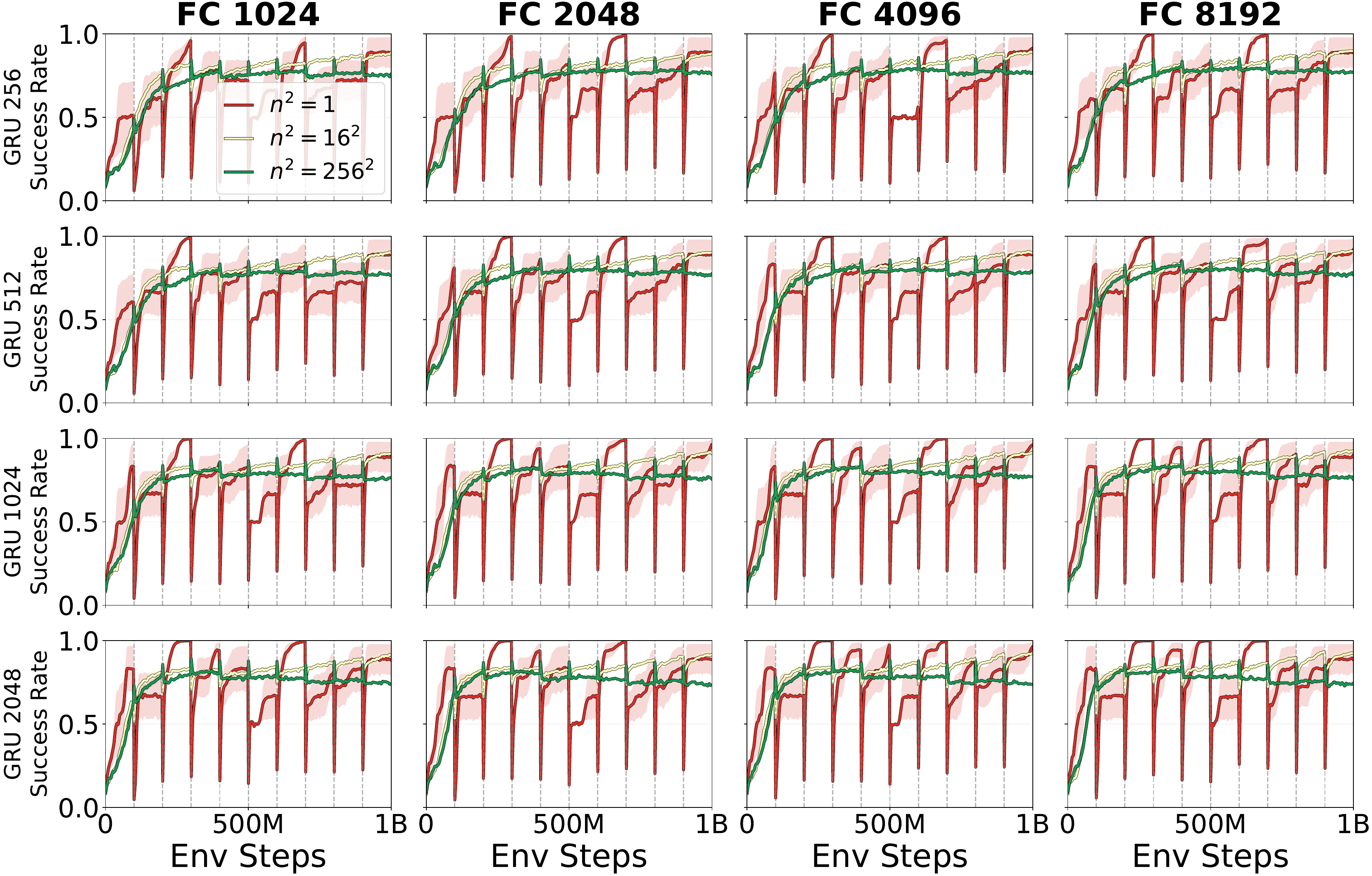}
    \caption{\textbf{Changing network capacity does not mitigate the negative effects of task diversity.} PPO + RNN across the ten-distribution sequence for every combination of GRU width $\{256, 512, 1024, 2048\}$ (rows) and actor and critic hidden width $\{1024, 2048, 4096, 8192\}$ (columns), at $n \in \{1, 16, 256\}$ with $3$ seeds each. Curves show success averaged over depths $1$--$6$; bands are standard error across seeds. At every size, the highest diversity produces the least learning over the sequence.}
    \label{fig:netsize}
\end{figure}

\figref{fig:netsize} repeats the ten-distribution PPO + RNN experiment across network sizes.

\end{document}